\documentclass{article} 
\usepackage{iclr2021_conference,times}
\usepackage{graphicx}
\usepackage{comment}
\usepackage{amsmath,amssymb} 
\usepackage{color}
\usepackage{subcaption}
\usepackage{enumitem}
\usepackage{booktabs}

\usepackage{amsmath,amsfonts,bm}

\def\eqref#1{equation~\ref{#1}}

\def\1{\bm{1}}

\DeclareMathAlphabet{\mathsfit}{\encodingdefault}{\sfdefault}{m}{sl}
\SetMathAlphabet{\mathsfit}{bold}{\encodingdefault}{\sfdefault}{bx}{n}

\usepackage{hyperref}
\usepackage{url}

\usepackage{diagbox}
\usepackage{verbatim}
\usepackage{multirow}
\usepackage{makecell}
\usepackage{pifont}
\newcommand{\cmark}{\ding{51}}%
\newcommand{\xmark}{\ding{55}}%

\definecolor{mybrown}{RGB}{165, 42, 42}
\usepackage{umoline}

\title{Does enhanced shape bias improve neural \\ network robustness to common corruptions?}

\author{Chaithanya Kumar Mummadi \thanks{Equal contribution.} \\
University of Freiburg\\
Bosch Center for Artificial Intelligence\\
\texttt{ChaithanyaKumar.Mummadi@bosch.com} \\
\And
Ranjitha Subramaniam \footnotemark[1]\\
Department of Computer Science \\
TU Chemnitz \\
\texttt{ranjivishnu08@gmail.com} \\
\And
Robin Hutmacher \\
Bosch Center for Artificial Intelligence\\
\texttt{Robin.Hutmacher@de.bosch.com} \\
\And
Julien Vitay \\
Department of Computer Science \\
TU Chemnitz \\
\texttt{julien.vitay@informatik.tu-chemnitz.de} \\
\And
Volker Fischer \\
Bosch Center for Artificial Intelligence\\
\texttt{Volker.Fischer@de.bosch.com} \\
\And
Jan Hendrik Metzen \\
Bosch Center for Artificial Intelligence\\
\texttt{JanHendrik.Metzen@de.bosch.com} \\
}

\iclrfinalcopy 
\begin{document}
\maketitle

\begin{abstract}
Convolutional neural networks (CNNs) learn to extract representations of complex features, such as object shapes and textures to solve image recognition tasks. 
Recent work indicates that CNNs trained on ImageNet are biased towards features that encode textures and that these alone are sufficient to generalize to unseen test data from the same distribution as the training data but often fail to generalize to out-of-distribution data. 
It has been shown that augmenting the training data with different image styles decreases this texture bias in favor of increased shape bias while at the same time improving robustness to common corruptions, such as noise and blur. 
Commonly, this is interpreted as shape bias increasing corruption robustness. However, this relationship is only hypothesized. We perform a systematic study of different ways of composing inputs based on natural images, explicit edge information, and stylization. While stylization is essential for achieving high corruption robustness, we do not find a clear correlation between shape bias and robustness. We conclude that the data augmentation caused by style-variation  accounts for the improved corruption robustness and increased shape bias is only a byproduct. 
\end{abstract}
\section{Introduction}
\label{sec:intro}
\begin{figure}[h]
	\begin{center}
		\includegraphics[width=0.92\linewidth, height=0.2\textheight]{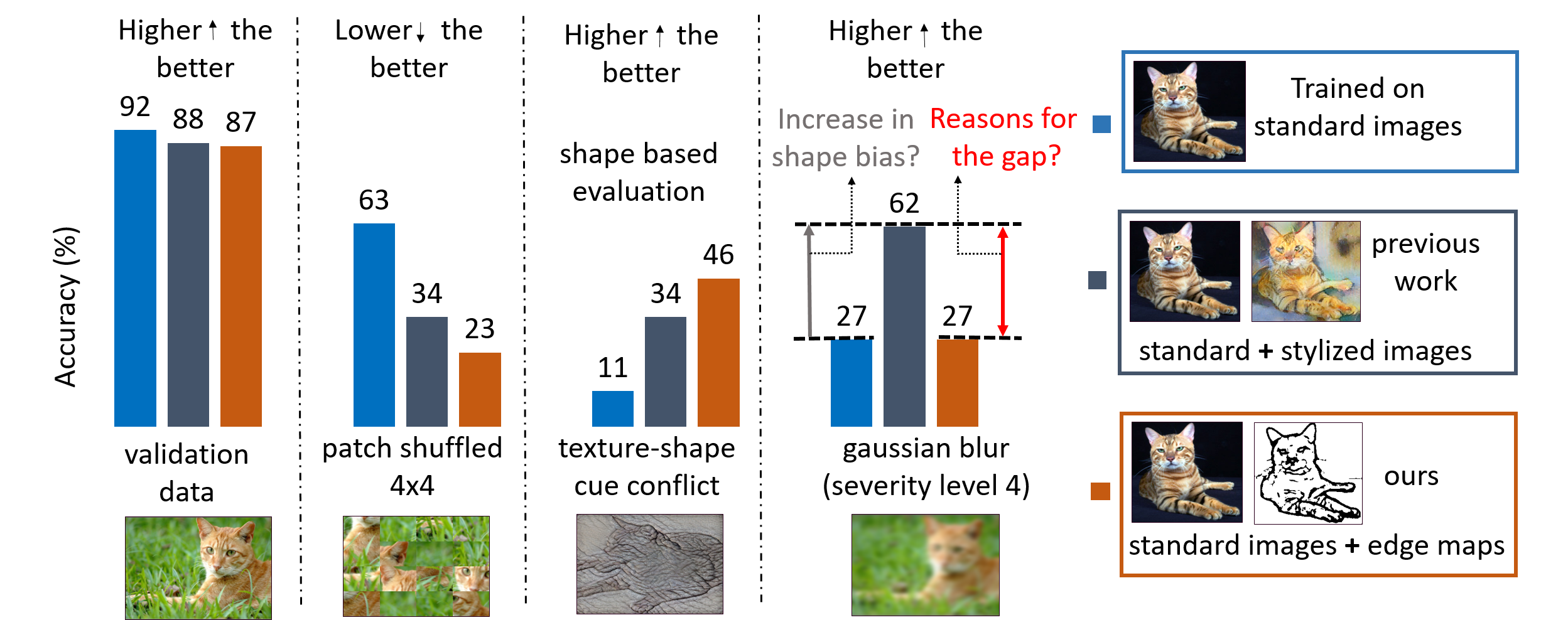}
		\caption{Illustration of the effect of different training augmentations. While both style-based \citep{geirhos2018imagenettrained} and edge-based augmentation (this paper) reach the same validation accuracy, edge-based augmentation shows a stronger increase in shape bias as evidenced by lower accuracy on patch-shuffled images and higher rate of classifying according to the shape category for texture-shape cue conflicts. Nevertheless, only style-based augmentation shows a considerable improvement against common corruptions such as Gaussian blur. This challenges the hypothesis that increased shape bias causes improved robustness to corruption.}
		\vskip -0.15in
		\label{fig:teaser}
	\end{center}
\end{figure}
As deep learning is increasingly applied to open-world perception problems in safety-critical domains such as robotics and autonomous driving, its robustness properties become of paramount importance. Generally, a lack of robustness against adversarial examples has been observed \citep{szegedy_intriguing_2013,goodfellow_explaining_2015}, making physical-world adversarial attacks on perception systems feasible \citep{kurakin_adversarial_2016,eykholt_physical_2018,lee_2019}. In this work, we focus on a different kind of robustness: namely, robustness against naturally occurring common image corruptions. Robustness of image classifiers against such corruptions can be evaluated using the ImageNet-C benchmark \citep{hendrycks_benchmarking_2018}, in which corruptions such as noise, blur, weather effects, and digital image transformations are simulated. \citet{hendrycks_benchmarking_2018} observed that recent advances  in neural architectures increased performance on undistorted data without significant increase in relative corruption robustness.

One hypothesis for the lack of robustness is an over-reliance on non-robust features that generalize well within the distribution used for training but fail to generalize to out-of-distribution data. \citet{ilyas_adversarial_2019} provide evidence for this hypothesis on adversarial examples. Similarly, it has been hypothesized that models which rely strongly on texture information are more vulnerable to common corruptions than models based on features encoding shape information \citep{geirhos2018imagenettrained,hendrycks_benchmarking_2018}. Alternative methods for increasing corruption robustness not motivated by enhancing shape bias use more (potentially unlabeled) training data \citep{xie2019self} or use stronger data augmentation  \citep{lopes2019improving,hendrycks2019augmix}. 
Note that our meaning of ``shape'' \& ``texture'' is built on the definitions by \citet{geirhos2018imagenettrained}.

In this paper, we re-examine the question of whether increasing the shape bias of a model actually helps in terms of corruption robustness. While prior work has found that there are training methods that increase both shape bias and corruption robustness \citep{geirhos2018imagenettrained,hendrycks_benchmarking_2018}, this only establishes a correlation and not a causal relationship. To increase the shape bias, \citet{geirhos2018imagenettrained} ``stylize'' images by imposing the style of a painting onto the image, leaving the shape-related structure of the image mostly unchanged while modifying texture cues so that they get largely uninformative of the class. Note that image stylization can be interpreted as a specific form of data augmentation, providing an alternative hypothesis for increased corruption robustness which would leave the enhanced shape bias as a mostly unrelated byproduct.

In this work, we investigate the role of the shape bias for corruption robustness in more detail. We propose two novel methods for increasing the shape bias:

 \begin{itemize}[noitemsep,topsep=0pt,parsep=0pt,partopsep=0pt, leftmargin=*]
	\item Similar to \citet{geirhos2018imagenettrained}, we pre-train the CNN on an auxiliary dataset which encourages learning shape features. In contrast to \citet{geirhos2018imagenettrained} that use stylized images, this dataset consists of the \emph{edge maps} for the training images that are generated using the pre-trained neural network of  \citet{liu2017richer} for edge detection. This method maintains global object shapes but removes texture-related information, thereby encouraging learning shape-based representations.
	\item In addition to pre-training on edge maps, we also propose \emph{style randomization} to further enhance the shape bias. Style randomization is based upon sampling parameters of the affine transformations of normalization layers for each input from a uniform distribution.
\end{itemize}

\begin{figure}[t]
\centering
\includegraphics[width=\linewidth]{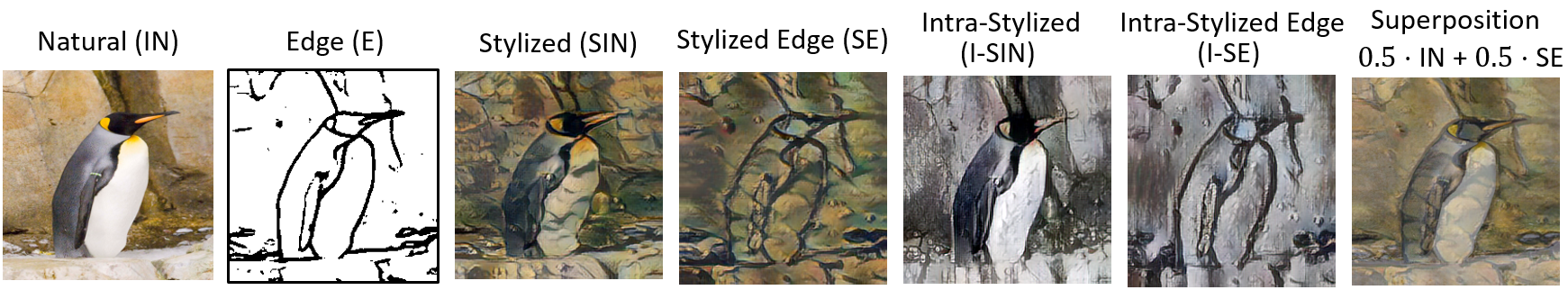}
\caption{Overview of content and stylization variants used in this paper: Content is a natural image (IN) or an edge map (E).
Content is stylized in three different ways: No stylization, style transfer with \texttt{Painter by Numbers} as style source as proposed in \cite{geirhos2018imagenettrained} (SIN and SE), style transfer with a different in-distribution image as style source (I-SIN and I-SE). Additionally, we show a superposition (SE+IN) between natural (IN) and stylized edge image (SE).}
\label{fig:teaser_dataset}
\end{figure}

Our key finding is summarized in Figure \ref{fig:teaser}. While pre-training on stylized images increases both shape bias and corruption robustness, these two quantities are not necessarily correlated: pre-training on edge maps increases the shape bias without consistently helping in terms of corruption robustness. In order to explain this finding, we conduct a systematic study in which we create inputs based on natural images, explicit edge information, and different ways of stylization (see Figure \ref{fig:teaser_dataset} for an illustration). 
We find that the shape bias gets maximized when combining edge information with stylization without including any texture information (Stylized Edges). However, for maximal corruption robustness, superimposing the image (and thus its textures) on these stylized edges is required. 
This, however, strongly reduces shape bias. In summary, corruption robustness seems to benefit most from style variation in the vicinity of the image manifold, while shape bias is mostly unrelated. Thus, image stylization is best interpreted as a strong data augmentation technique that encourages robust representations, regardless whether these representations are shape-based or not. 

Moreover, we present results for a setting where we fine-tune only parameters of the affine transformation of a normalization layer on the target distribution (stylized or corrupted images, respectively) for a CNN trained on regular images. Surprisingly, this is already sufficient for increasing the shape bias/corruption robustness considerably. We conclude that CNNs trained on normal images do learn shape-based features and features robust to corruptions but assign little weight to them. It may thus be sufficient to perform augmentation in feature space (extending \citet{nam2019reducing,li2020feature}) so that higher weights are assigned to features that are robust to relevant domain shifts.
\section{Related work}
\label{sec:related_work}

\textbf{Texture-vs-Shape Bias}
\citet{geirhos2018imagenettrained} and \citet{baker2018deep} hypothesized that CNNs tend to be biased towards textural cues rather than shape cues. 
This line of research is further supported by \citet{brendel2018approximating}, where the authors show that BagNets, Deep Neural Networks (DNN) trained and evaluated only on small restricted local image patches, already perform reasonably well on ImageNet. 
Similarly, \citet{yin2019fourier} and \citet{jo2017measuring} showed using a Fourier space analysis that DNNs rely on surface statistical regularities and high-frequency components. 
The texture-vs-shape bias can be quantified by evaluating a network either on images with texture-shape cue conflict \citep{geirhos2018imagenettrained} or on images which were patch-wise shuffled \citep{luo2019defective}.

\textbf{Robustness Against Common Corruptions}
Common corruptions are potentially stochastic image transformations motivated by real-world effects that can be used for evaluating model robustness. 
\citet{hendrycks_benchmarking_2018} proposed the ImageNet-C dataset that contains simulated corruptions such as noise, blur, weather effects and digital image transformations. 
\citet{geirhos2018generalisation} showed that humans are more robust to image corruptions than CNNs.

Approaches to improve corruption robustness include data augmentation \citep{lopes2019improving,yun2019cutmix,hendrycks2019augmix,cubuk2019randaugment}, self-training with more training data \citep{xie2019self}, novel architectures and building blocks \citep{zhang2019shiftinvar,hu2018squeeze}, and changes in the training procedure \citep{hendrycks2019using,Rusak2020a,NIPS2019_9237}. 
Motivated by the texture-vs-shape hypothesis, \citet{geirhos2018imagenettrained} and  \citet{michaelis2019benchmarking} train their network on a stylized version of ImageNet. 
The idea is that style transfer removes textural cues and models trained on stylized data thus have to rely more on shape information.
The observed increase in corruption robustness on this stylized data was attributed to the shape bias.
In this work, we provide evidence that contradicts this claim.

Similar to training on stylized images, \emph{Style Blending} \citep{nam2019reducing} employs style transfer in latent space by interpolating between feature statistics of different samples in a batch. 
\citet{li2020feature} extend this idea and use feature space blending along with label interpolation. \citet{hendrycks2019using} considers self-supervised training with the prediction of image rotations as an auxiliary task. 
The authors argue that predicting rotation requires shape information and thus improves robustness.
Similarly, \citet{shi2020informative} proposes Dropout-like algorithm to reduce the texture bias and thereby increase the shape bias to improve model robustness. However, the authors also discuss that a ``sweet spot" between shape and texture is needed for the model to be robust for domain generalization.
With \emph{Patch-wise Adversarial Regularization}, \citet{NIPS2019_9237} try to penalize reliance on local predictive representations in early layers and encourage the network to learn global concepts. 
Other augmentation techniques that aim to improve common corruption robustness are \emph{PatchGaussian} \citep{lopes2019improving}, \emph{CutMix} \citep{yun2019cutmix}, \emph{AugMix} \citep{hendrycks2019augmix}, and \emph{RandAugment} \citep{cubuk2019randaugment}. 
At this point, it remains unclear whether the increase in robustness caused by these augmentations is due to learning fundamentally different representations such as more shape-biased ones or to more incremental improvements in feature quality.

\textbf{Edge-based Representations}
A classical method for extracting edge maps is the Canny edge extractor \cite{canny1986computational}. 
More recent approaches use DNNs \citep{xie2015holistically,liu2017richer} (see Figure \ref{fig:edgemaps}). 
\citet{geirhos2018imagenettrained} evaluate their shape-biased models on edge maps obtained with a Canny edge detector. 
ImageNet-Sketch \citep{NIPS2019_9237} is a newly collected sketch-like dataset matching the ImageNet validation dataset in shape and size. It is used to evaluate generalization to domain shifts. In contrast to these works, we generate the edge-based representations with an edge detector using Richer Convolutional Features (RCF) \citep{liu2017richer} (see Figure \ref{fig:edgemaps}) and use them explicitly for training. We provide evidence that edge-based representations enhance the shape bias, through an evaluation on images with induced texture-shape cue conflict and patch-shuffled images.

\section{Learning shape-based representations}
\label{sec:approach}

Similar to \citet{geirhos2018imagenettrained}, we aim to enhance the shape bias of a network so that it bases its decision more on shape details than on the style of objects encoded in textures. 
While \citet{geirhos2018imagenettrained} augment training data with different styles (stylization), thereby making texture cues less predictive,  we extract edge information (edge maps) from the training images to maintain explicit shape details and remove texture-related information completely. Here, we consider grayscale intensity edge maps rather than separate edge maps for each color channel. We propose to train CNNs using the edge maps in addition to the standard training data to learn shape-based representations for more effective shape-based decision-making.

Besides training on the dataset with explicit shape cues, high capacity networks learn different feature representations when trained jointly on datasets from different distributions. 
Despite edge maps encouraging CNNs to learn shape-based representations, we observe that the network learns to encode features with texture details when introduced to the standard image data during training. 
We propose here to further restrain the network from learning texture details on standard image data. 
We discuss below the extraction of edge details from images to create the edge map dataset and explain the technique to reduce the texture bias of the CNN.

\textbf{Edge dataset}
Given a standard image dataset, we construct a new dataset with edge maps (named the Edge dataset) by extracting the edge details of each image. The edge details are extracted by the CNN-based edge detector using richer convolutional features (RCF) proposed in \cite{liu2017richer}. RCF network produces a single-channel edge map that contains the pixel values between $[0,\,255]$. 
We convert the non-binary edge map into a binary map with values in \(\{0, 255\}\) using a threshold of \(128\) and transform it into a \(3\)-channel RGB edge map by duplicating the channels, so we can use the edge maps as a direct input to train the CNNs. The edge maps from the Edge dataset are used as input and can be independently used to train or evaluate CNNs without necessarily being combined with the standard image data. Please refer to Section \ref{app_sec:edge_dataset} for the details of RCF network.

\textbf{Style Randomization (\emph{SR})}
While using a dataset with explicit shape cues enhances shape-based representations, we propose to further reduce the texture bias of the network when training on standard images. 
It is shown in the literature of style transfer \citep{dumoulin2016learned,huang2017arbitrary} that the statistics of feature maps (e.g., mean and standard deviation) of a CNN effectively capture the style of an image and changing these statistics would correspond to a change in the style of an image. SIN dataset is generated using such style transfer technique and shown to reduce the texture bias of the networks. Inspired by this observation, we propose a simple technique to effectively reduce the texture bias using the feature statistics when being trained on standard training data. We modify the style of an image in the feature space so that the network becomes style-invariant. In particular, we randomize the style details, i.e. feature statistics, of an image during training such that the network can not rely on the texture cues. A similar approach named \emph{Style Blending} (SB) is proposed in \citet{nam2019reducing} which randomizes the style information by interpolating the feature statistics between different samples in a mini-batch. We propose here a slightly different approach to make the network invariant to style information.
Instead of interpolating the statistics of similar distribution of data i.e, training samples, we completely randomize the feature statistics (mean and standard deviation) by randomly sampling them from an uniform distribution. 
Considering $X_i$ as the $i^{th}$ feature map of an intermediate layer in CNN, and $\mu_i$ \& $\sigma_i$ as the feature statistics of $X_i$, the style randomized feature map $\hat X_i$ is defined as:

\begin{equation}
\hat X_i := \hat\sigma_i * \left(\frac{X_i - \mu_i}{\sigma_i}\right) + \hat\mu_i
\end{equation}

where $\hat\sigma_i \sim$ Uniform$(0.1, 1)$  and $\hat\mu_i \sim$ Uniform$(-1, 1)$. 
These specific choices of sampling for $\hat\sigma_i$ and $\hat\mu_i$ were found to perform best on our evaluations.
The style transfer technique described in \cite{huang2017arbitrary} replaces the feature statistics of content image with the statistics of a desired style image to change the style.
Similarly, we replace the statistics of content image with random statistics to change the style information.
Training the network with \emph{SR} reduces the texture bias and improves shape-based decision making. 
An advantage of \emph{SR} over \emph{SB} is that the feature statistics are sampled from a different distribution than the training data, that encourages learning representations to generalize better to out-of-distribution data. We show in Section \ref{sec:evaluation} that \emph{SR} outperforms \emph{SB} and aids the network to induce stronger shape-based representations.

\section{Experimental settings}
\label{sec:experiments}

\textbf{Dataset}
We use a subset of \(20\) classes from ImageNet dataset (ImageNet20, or IN) that are chosen randomly, to study the role of shape bias towards corruption robustness; the main reason being that extensive experiments on this dataset are feasible with limited computation. Details about this dataset can be found in Section \ref{app_sec:IN_dataset}.
The Edge dataset of IN (referred to as E) is generated as described in Section \ref{sec:approach}. 
 
\textbf{Stylization variants}
In addition to enhancing the shape bias using the edge maps,
we further study the contribution of different factors of Stylized ImageNet (SIN) \citep{geirhos2018imagenettrained} to gain insights on its improved performance on corruptions. We break down SIN into different factors to understand their influence on corruption robustness. We segregate the factors that jointly generate the stylized images and the factors that are hypothesized to improve corruption robustness. These include i) shape bias of the network, ii) styles that are transferred from paintings and iii) statistics of natural images from IN. The \emph{role of shape bias} is studied using the Edge dataset (E) proposed in Section \ref{sec:approach}. Other variants study the role of the remaining factors and are explained below:

\emph{Role of stylization} We create Stylized Edges (SE, see Figure \ref{fig:teaser_dataset}) for which the styles from the paintings are transferred to the edge maps of Edge ImageNet20 (E). Here, we study the significance of stylization without the presence of the statistics (texture details) of natural images.

\emph{Role of out-of-distribution styles} SIN is generated by transferring the styles from out-of-distribution images, namely paintings. We create its variant called Intra-Stylized IN (I-SIN, see Figure \ref{fig:teaser_dataset}) for which in-distribution images from IN are chosen randomly to transfer the styles. We also generate Intra-Stylized Edges (I-SE) where the image styles of IN are transferred to the Edge dataset E.

\emph{Role of natural images statistics} 
The above variants of E or SE test the role of shape and stylization without retaining texture cues of natural images. We create another variant called \emph{Superposition} (SE+IN, see Figure \ref{fig:teaser_dataset}) that interpolates images \(I_{\text{SE}}\) from SE with images \(I_{\text{IN}}\) from IN to embed the statistics (texture details) from natural images: $I_{\text{SE+IN}} := (1-\alpha)\cdot I_{\text{SE}} + \alpha\cdot I_{\text{IN}}$. We set $\alpha=0.5$. 

These different stylized variants including E allow insights into the interplay between shape bias and corruption robustness. 
For simplicity, we term the networks that are trained on a certain dataset using the name of that dataset. For example, network trained on Stylized Edge (SE) is referred to as SE. The evaluation of SIN and I-SIN reveals the significance of the choice of styles and evaluation of Edge (E) indicates the role of the shape bias for corruption robustness. SE explains the importance of stylization and finally SE+IN allows to understand the importance of natural image statistics that are preserved in SIN and I-SIN but are missing from SE. Table \ref{tab:finetune_corruptions} provides an overview of the input image compositions of different variants that are described above.



\textbf{Network details}
We employ a ResNet18 architecture with group normalization \citep{wu2018group} and weight standardization \citep{qiao2019weight}.
We include \emph{SR} described in Section \ref{sec:approach} in the architecture. ResNet18 contains 4 stages of series of residual blocks and \emph{SR} is inserted before every stage. 
We train ResNet18 on different datasets and their variants described above. IN and SIN are considered as baselines. We show that E possesses more global shape details of the objects whereas SIN demonstrates little or no texture bias for decision making. Both these datasets are complementary to each other and further enhance shape-based predictions when combined (termed as E-SIN). 
Note that \emph{SR} is used to reduce texture bias and IN contains by far the strongest texture cues. Hence,  \emph{SR} is applied only on the training samples of IN but not on other dataset variants. Nevertheless, \emph{SR} applied on other dataset variants found no differences in the results.

\textbf{Training details}
Network on differnt dataset variants except IN are trained in two stages. The first stage begins with training the network on the respective dataset variant (e.g: E) for a total of 75 epochs starting with a learning rate of \(0.1\), which is dropped at the 60th epoch by a factor 10. In the second stage, the networks are then fine-tuned on the respective dataset along with IN (e.g: E \& IN) for another round of 75 epochs starting with a learning rate of \(0.01\), later reduced to \(0.001\) at the 60th epoch. On the other hand, the network on IN is trained for 100 epochs with a learning rate of \(0.1\), reduced to \(0.01\) and \(0.001\) at the 60th and 90th epochs, respectively. We use a batch size of \(128\) samples with the SGD optimizer and weight decay \(10^{-4}\).

During the fine-tuning stage, we freeze the first convolutional layer and the first normalization layer's affine parameters. We observed that freezing these two layers demonstrate more global shape bias than fine-tuning all the layers in the network. During fine-tuning, the networks receive an equal number of training samples from both datasets (e.g: 128 samples from E and 128 samples from IN in a mini-batch). Note that the data distribution of edge maps from the datasets E, SE and I-SE are different than the distribution of images from other datasets. Fine-tuning the network on inputs with different distributions results in degradation of the performance. In other words, the datasets E, SE and I-SE do not preserve natural image statistics and degrade task performance when finetuning along with clean images. Hence, we weigh the loss of training samples of edge maps from E, SE and I-SE when fine-tuning along with IN. The loss between training samples is weighted as follows: $Loss \ L$ = $L_{\text{IN}} + \lambda L_{\text{edgemaps}}$, with $\lambda=0.01$. Finetuning on style variants SIN, I-SIN that better preserve natural image statistics does not affect classification performance significantly, hence $\lambda$ is not used. Larger $\lambda$ preserves the shape bias but affects the clean accuracy while smaller $\lambda$ reduces shape bias of the network.
In case of E-SIN,
We fine-tune the network that is pre-trained on E in the first stage of training with SIN and IN in the second stage and show that such setup further improves shape-based predictions.
All ResNet18 models have validation accuracy of about 87\% on IN.

\section{Evaluation of shape bias}
\label{sec:evaluation}
\begin{table}[t]
\centering
\begin{tabular}{|l||l|l|l||l|l|l|}
\hline
 & \multicolumn{3}{c||}{\thead{shuffled image patches $4 \times 4$ acc(\%)}}  & \multicolumn{3}{c|}{\thead{shape based cue conflict \#400}}\\  \cline{2-7} 
\thead{Network\\} &   \thead{No \\styling} & \thead{style \\blending} & \thead{style \\randomization} & \thead{No \\styling} & \thead{style \\blending} & \thead{style \\randomization}\\ \hline
\makecell{IN}   & \makecell{67.22}  & \makecell{51.34}  & \makecell{41.97} & \makecell{63}  & \makecell{82}  & \makecell{86}  
\\ \hline
\makecell{SIN} & \makecell{38.46}  & \makecell{36.96}   & \makecell{34.95} & \makecell{144}     & \makecell{155} & \makecell{156}               
\\ \hline
\makecell{E}  & \makecell{34.11}    & \makecell{33.95}   & \makecell{\textbf{28.43}} & \makecell{155}    & \makecell{166}  & \makecell{\textbf{193}}
\\ \hline

\end{tabular}
\caption{Comparison of different feature space style augmentation methods on $4 \times 4$ shuffled image patches and number of shape based predictions in texture-shape cue conflict images. Evaluation of shuffled patches is conducted on \(598\) correctly classified validation images by all the networks.}
	\vskip -0.15in
	\label{tab:stylization}
\end{table}


\begin{table}[t]
\centering
\begin{tabular}{|l||l|l|l||l|l|l|}
\hline
 & \multicolumn{3}{c||}{\thead{shuffled image patches acc(\%)}}  & \multicolumn{3}{c|}{\thead{texture-shape cue conflict results}}\\  \cline{2-7} 
\thead{Network\\} &   \thead{\textbf{$2 \times 2$}}  & \thead{$4 \times 4$}  & \thead{$8 \times 8$} & \thead{shape \#400} & \thead{shape \#100} & \thead{texture \#100}\\ \hline
\makecell{IN}   & \makecell{78.57}  & \makecell{41.93}  & \makecell{31.21} & \makecell{86}  & \makecell{18}  & \makecell{20}  
\\ \hline
\makecell{SIN} & \makecell{75.78}  & \makecell{35.56}   & \makecell{18.48} & \makecell{156}     & \makecell{32} & \makecell{\bf 2}               
\\ \hline
\makecell{E}  & \makecell{73.29}    & \makecell{28.42}   & \makecell{11.18} & \makecell{193}    & \makecell{46}  & \makecell{15}  
\\ \hline
\makecell{SE}  & \makecell{\bf 66.77}  & \makecell{28.73} & \makecell{12.89}  & \makecell{224}     &  \makecell{55} & \makecell{6}    
\\ \hline
\makecell{E-SIN} & \makecell{71.12} & \makecell{\bf 23.76} & \makecell{\bf 10.25} & \makecell{\bf 234}   & \makecell{\bf 58}  & \makecell{6}        
\\ \hline

\end{tabular}
\caption{Comparison of models trained on different datasets on shuffled image patches and number of texture-shape cue conflict predictions based on shape and texture labels. Evaluation of shuffled image patches is conducted on \(644\) validation images that are correctly classified by all the networks.}
	\vskip -0.15in
	\label{tab:shape_bias}
\end{table}

In this section, we evaluate different methods in terms of their shape bias using two different evaluation criteria - \emph{Shuffled image patches} and \emph{Texture-shape cue conflict} that are described below.

\textbf{Shuffled image patches:}
Following \citet{luo2019defective}, we manipulate images by perturbing the shape details while preserving the local texture of the objects. We divide an image into different patches of size $n\times n$ with $n\in \{2,4,8\}$ and randomly shuffle the patches as shown in Figure \ref{fig:shuffled_patches}. Larger $n$ corresponds to more distorted shapes. The performance of networks that rely more on shape is expected to deteriorate more strongly as the number of patches increases. 
We conduct this evaluation only on the ImageNet20 validation images that were correctly classified by \emph{all} the networks that are selected for comparison. 

\textbf{Texture-shape cue conflict:}
The cue conflict image dataset proposed by \citet{geirhos2018imagenettrained} consists of images where the shape of an object carries the texture of a different object. For example, the object \texttt{cat} holds the texture of \texttt{elephant} as shown in Figure \ref{fig:cue_conflict}. Each image in the dataset contains two class labels: labels with respect to shape and texture. The evaluation is carried out to test the network's bias towards shape or texture. Networks with strong shape bias will exhibit higher accuracy according to the shape label while networks with texture bias will have higher accuracy for texture-based label. The original dataset contains a total of \(1280\) cue conflict images designed for the evaluation of the networks trained on the entire ImageNet dataset. \(400\) of these images have classes (shape labels) present in ImageNet20. A subset of \(100\) instances (\(20\) instances from \(5\) different categories) from the selected images also has a texture label that belongs to ImageNet20 (see Figure \ref{fig:cue_conflict} bottom). The remaining \(300\) images with texture labels that do not belong to the classes of ImageNet20 are not considered for texture-based classification. 

\textbf{Results}
The results in Table \ref{tab:stylization} compare \emph{style blending (SB)} \citep{nam2019reducing}, \emph{style randomization (SR)} (Section \ref{sec:approach}), and no styling in feature space for networks trained on IN, SIN and E. In terms of performance on $4\times 4$ shuffled patches, \emph{SB} performs worse than no styling, and \emph{SR} performs even worse than \emph{SB}. This indicates increasing shape bias from no styling over \emph{SB} to \emph{SR}. This finding is reinforced by an increasing number of images classified according to the shape label for texture-shape cue conflict images from no styling over \emph{SB} to \emph{SR}. Similarly, when comparing different training datasets, SIN results in stronger shape bias than IN, and E exhibits stronger shape bias than SIN. 

In Table \ref{tab:shape_bias}, we compare additional networks, all with \emph{SR} enabled. Here, we again see a consistent trend of increasing shape bias from IN over SIN to E. Moreover, stylized edges (SE) further increase shape bias than E. Lastly, E-SIN improves shape bias even slightly beyond SE. In summary, we can see a clear increase in shape bias for the methods proposed in this paper over IN or SIN. Next, we investigate to which extent this also results in an increased corruption robustness.

\section{Influence of shape bias on common corruptions}
\label{sec:robustness_corruptions}
\begin{figure}[t]
	\begin{center}
		\includegraphics[width=1.0\linewidth, height=0.25\textheight]{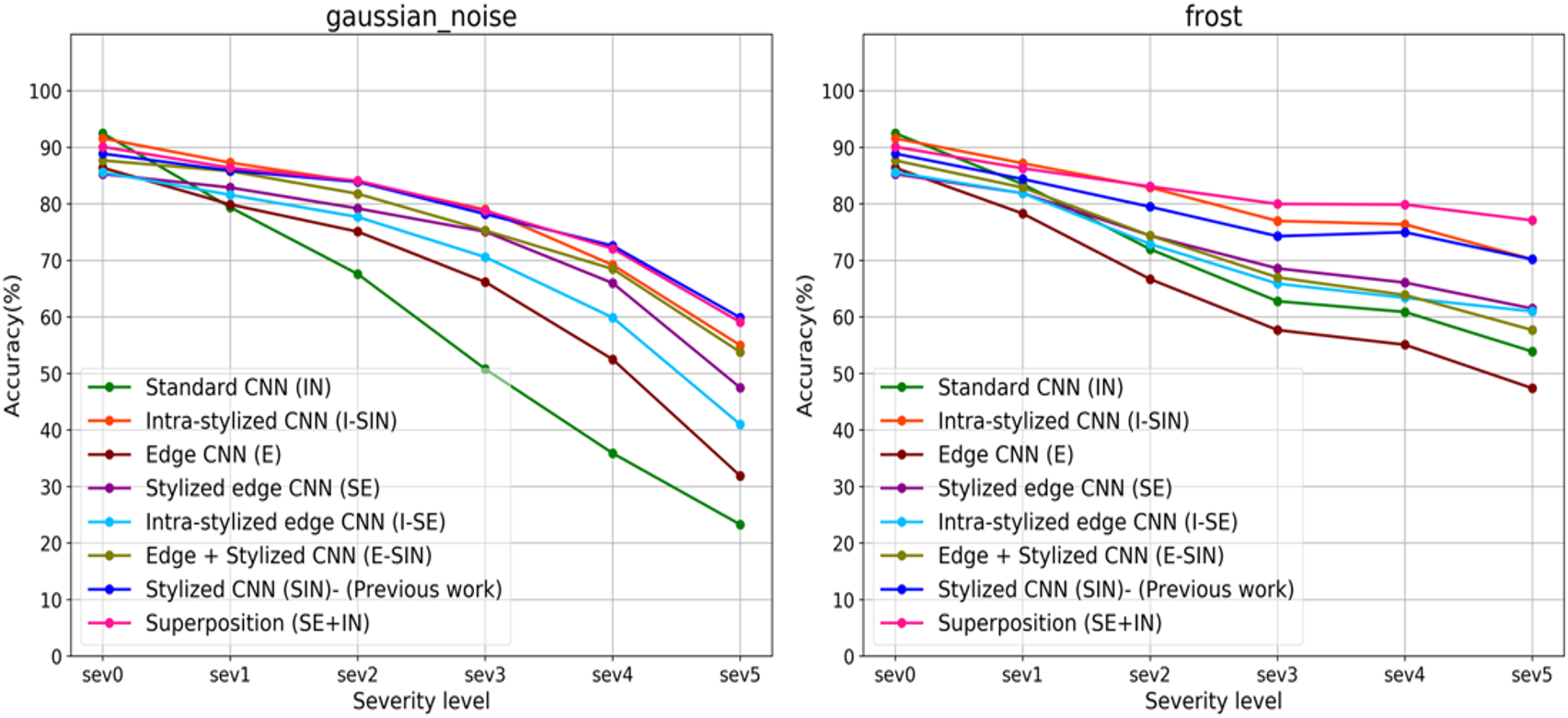}
		\caption{Classification accuracy of different networks on two corruptions across \(5\)  severity levels. Severity \(0\) represents accuracy on clean validation data of IN. Severity levels \(1\) - \(5\) follow the corruption parameters from \citet{hendrycks_benchmarking_2018} and represent increasingly strong corruptions.}
		\vskip -0.15in
		\label{fig:patch_shuffle_results}
	\end{center}
\end{figure}

\begin{table}[tb]
\centering
\begin{tabular}{|l||l|l|l||l|l|l|}
\hline
 & \multicolumn{3}{c||}{\thead{Input image composition}}  &          &     &  \\  \cline{2-4} 
\thead{Network\\} &   \thead{Natural \\image} & \thead{Edge\\ map} & \thead{Style\\ transfer} & \thead{Shape\\ \#100} & \thead{Texture \\ \#100} & \thead{Mean corruption \\acc(\%)}\\ \hline
\makecell{IN}   & \makecell{\cmark}  & \makecell{\xmark}  & \makecell{\xmark} & \makecell{11}  & \makecell{39}  & \makecell{64.69}  
\\ \hline
\makecell{SIN} & \makecell{\cmark}  & \makecell{\xmark}   & \makecell{\cmark} & \makecell{34}     & \makecell{2} & \makecell{77.64}               
\\ \hline
\makecell{E}  & \makecell{\xmark}    & \makecell{\cmark}   & \makecell{\xmark} & \makecell{46}    & \makecell{15}  & \makecell{62.01}  
\\ \hline
\makecell{SE}  & \makecell{\xmark}  & \makecell{\cmark} & \makecell{\cmark}  & \makecell{55}     &  \makecell{6} & \makecell{71.81}    
\\ \hline
\makecell{E-SIN} & \makecell{\cmark} & \makecell{\cmark} & \makecell{\cmark} & \makecell{62}   & \makecell{5}  & \makecell{71.55}    
\\ \hline
\makecell{SE+IN} & \makecell{\cmark} & \makecell{\cmark} & \makecell{\cmark} & \makecell{22}   & \makecell{13}  & \makecell{\textbf{78.96}}          
\\ \hline

\end{tabular}
\caption{Mean corruption accuracy (mCA) and texture/shape results on texture-shape cue conflict dataset of different networks. mCA is the mean accuracy over \(15\) ImageNet-C corruption and severities ranging from \(1\) to \(5\). Networks trained with style transfer augmentation perform better than those without and network trained on superpositioned images (SE+IN) yield best mCA.}
	\vskip -0.15in
	\label{tab:finetune_corruptions}
\end{table}

We compare different networks in terms of their corruption robustness. Figure \ref{fig:patch_shuffle_results} shows the accuracy of different networks for two types of corruptions: Gaussian noise and frost (refer Figure \ref{fig:appendix_networks_corruptions} for all corruptions). Table \ref{tab:finetune_corruptions} presents the corruption accuracy averaged over 15 ImageNet-C corruptions along with shape and texture results on the texture-shape cue conflict dataset.
Generally, a CNN trained on IN performs poorly in terms of corruption robustness while SIN is relatively robust. On the other hand, E performs considerably worse than SIN and is not consistently better than IN despite having an even stronger shape bias than SIN. Networks SE and E-SIN further increase shape bias but still perform considerably worse than SIN in terms of corruption robustness. These results contradict the hypothesis that stronger shape bias results in increased corruption robustness. 

The only method that slightly surpasses SIN in terms of corruption robustness is the superposition of SE with natural images (SE+IN). However, this method has a relatively small shape bias. A common theme of SIN and SE+IN is that both exhibit properties of a natural image but are strongly distorted by stylization (see Figure \ref{fig:teaser_dataset}). We hypothesize that these methods correspond to strong augmentation methods that stay close enough to the data manifold while inducing high diversity in appearance and thereby encourage learning robust representations, which need not necessarily be shape-based. 
We extend these findings to larger datasets with 200 classes of ImageNet, deeper architectures like ResNet50, DenseNet121, MobileNetV2  and different normalization layers like BatchNorm in Section \ref{app_sec:extended experiments}.
Lastly, as can be seen from Figure \ref{fig:patch_shuffle_results}, intra-stylization is nearly as effective as stylization based on paintings, implying that style need not necessarily be out-of-distribution for being useful. 
\section{On the adaptability of learned representations}
\label{sec:finetune_affine}
\begin{table}[tb]
\centering
\begin{tabular}{|l||l|l|l|l||l|l|l|}
\hline
 & \multicolumn{4}{c||}{\thead{Corruptions acc(\%)}} & & \multicolumn{2}{c|}{\thead{Cue Conflict}} \\
 \cline{2-5} \cline{7-8}

 \thead{Network}       &     \thead{Speckle\\ noise} & \thead{Gaussian\\ blur} & \thead{Frost} & \thead{Pixelate}  & \thead{SIN \\val acc(\%)}  &  \thead{shape\\ \#400} & \thead{texture \\ \#100}\\  \hline
\makecell{IN}                                                     & \makecell{61.28}         & \makecell{42.96}         & \makecell{66.62} & \makecell{78.54}        & \makecell{42.0}    & \makecell{63}  & \makecell{39}      \\ \hline
\makecell{IN (fine-tuned)} & \makecell{82.7}          & \makecell{77.3}          & \makecell{81.02} & \makecell{87.02} &    \makecell{68.0}   & \makecell{130}   & \makecell{13}    \\ \hline
\makecell{E}                                                        & \makecell{67.76}      & \makecell{44.48}         & \makecell{61.04} & \makecell{70.94}      &   \makecell{62.3}     & \makecell{193}  & \makecell{15}      \\ \hline
\makecell{E (fine-tuned)}    & \makecell{80.18}         & \makecell{71.74}         & \makecell{73.7}  & \makecell{74.78} & \makecell{72.4}    & \makecell{222}  & \makecell{9}              \\ \hline
\end{tabular}
\caption{Mean corruption accuracy, SIN and cue conflict results of networks with \& without additional fine-tuning of the affine parameters of normalization layers on the respective corruptions. Fine-tuned networks perform significantly better, despite only the normalization layers are updated.}
	\vskip -0.15in
	\label{tab:finetune_corruptions_sin}
\end{table}
As seen in the previous section, style augmentation on natural images is important for the network to be able to generalize to different domains such as common corruptions. We now study how easily a pre-trained network can be adapted to a different distribution such as corruptions. Importantly, this uses the ``unknown'' distortion during training; this experiment is not meant as a practical procedure for the ImageNet-C benchmark but rather for understanding internal representations of a network.

\cite{chang2019domain} showed that domain-specific affine parameters in normalization layers are essential when training a network on different input data distributions jointly. We conduct a similar experiment with the key difference that our network is already pre-trained on IN/E and only the affine parameters of normalization layers are fine-tuned to fit the distribution of the respective target domain. First, we fine-tune affine parameters of the network on several ImageNet-C corruptions separately and evaluate the mean corruption accuracy on the same corruption across different severity levels. As shown in Table \ref{tab:finetune_corruptions_sin} (left), performance on the corruptions can be greatly improved even with fixed convolutional parameters trained on IN/E by just tuning the affine parameters. Similarly, we also fine-tune the affine parameters of pre-trained CNN on SIN. Results in Table \ref{tab:finetune_corruptions_sin} (right) show not only an improvement on SIN validation accuracy but also improved shape-based classification results on texture-shape cue conflict images. These results suggest that the standard CNN encodes robust representations that can be leveraged when adapting affine parameters on a target domain.
\section{Conclusion}
We performed a systematic empirical evaluation of the hypothesis that enhanced shape bias of a neural network is predictive for increased corruption robustness. Our evidence suggests that this is not the case and increased shape bias is mostly an unrelated byproduct. Increased corruption robustness by image stylization is better explained as a strong form of augmentation which encourages robust representations regardless whether those are shape-based or based on other cues. We conclude that if corruption robustness is the main objective, one should not primarily focus on increasing the shape bias of learned representations.
Potential future research directions will focus on understanding whether shape-biased representations offer advantages in other domains than corruption robustness \citep{hendrycks2020many}. Moreover, one could try devising stronger augmentation procedures in image or feature space based on our findings. Lastly, gaining a better understanding of which kind of features (if not shape-based ones) are essential for corruption robustness is an important direction.

\bibliography{iclr2021_conference}

\begin{thebibliography}{38}
\providecommand{\natexlab}[1]{#1}
\providecommand{\url}[1]{\texttt{#1}}
\expandafter\ifx\csname urlstyle\endcsname\relax
  \providecommand{\doi}[1]{doi: #1}\else
  \providecommand{\doi}{doi: \begingroup \urlstyle{rm}\Url}\fi

\bibitem[Arbelaez et~al.(2010)Arbelaez, Maire, Fowlkes, and
  Malik]{arbelaez2010contour}
Pablo Arbelaez, Michael Maire, Charless Fowlkes, and Jitendra Malik.
\newblock Contour detection and hierarchical image segmentation.
\newblock \emph{IEEE transactions on pattern analysis and machine
  intelligence}, 33\penalty0 (5):\penalty0 898--916, 2010.

\bibitem[Baker et~al.(2018)Baker, Lu, Erlikhman, and Kellman]{baker2018deep}
Nicholas Baker, Hongjing Lu, Gennady Erlikhman, and Philip~J Kellman.
\newblock Deep convolutional networks do not classify based on global object
  shape.
\newblock \emph{PLoS computational biology}, 14\penalty0 (12):\penalty0
  e1006613, 2018.

\bibitem[Brendel \& Bethge(2019)Brendel and Bethge]{brendel2018approximating}
W.~Brendel and M.~Bethge.
\newblock Approximating cnns with bag-of-local-features models works
  surprisingly well on imagenet.
\newblock May 2019.

\bibitem[Canny(1986)]{canny1986computational}
John Canny.
\newblock A computational approach to edge detection.
\newblock \emph{IEEE Transactions on pattern analysis and machine
  intelligence}, \penalty0 (6):\penalty0 679--698, 1986.

\bibitem[Chang et~al.(2019)Chang, You, Seo, Kwak, and Han]{chang2019domain}
Woong-Gi Chang, Tackgeun You, Seonguk Seo, Suha Kwak, and Bohyung Han.
\newblock Domain-specific batch normalization for unsupervised domain
  adaptation.
\newblock In \emph{Proceedings of the IEEE Conference on Computer Vision and
  Pattern Recognition}, pp.\  7354--7362, 2019.

\bibitem[Cubuk et~al.(2019)Cubuk, Zoph, Shlens, and Le]{cubuk2019randaugment}
Ekin~D Cubuk, Barret Zoph, Jonathon Shlens, and Quoc~V Le.
\newblock Randaugment: Practical automated data augmentation with a reduced
  search space.
\newblock \emph{arXiv preprint arXiv:1909.13719}, 2019.

\bibitem[Dumoulin et~al.(2016)Dumoulin, Shlens, and
  Kudlur]{dumoulin2016learned}
Vincent Dumoulin, Jonathon Shlens, and Manjunath Kudlur.
\newblock A learned representation for artistic style.
\newblock \emph{arXiv preprint arXiv:1610.07629}, 2016.

\bibitem[Eykholt et~al.(2018)Eykholt, Evtimov, Fernandes, Li, Rahmati, Tramer,
  Prakash, Kohno, and Song]{eykholt_physical_2018}
Kevin Eykholt, Ivan Evtimov, Earlence Fernandes, Bo~Li, Amir Rahmati, Florian
  Tramer, Atul Prakash, Tadayoshi Kohno, and Dawn Song.
\newblock Physical {Adversarial} {Examples} for {Object} {Detectors}.
\newblock In \emph{12th USENIX Workshop on Offensive Technologies}, July 2018.
\newblock URL \url{http://arxiv.org/abs/1807.07769}.

\bibitem[Geirhos et~al.(2018)Geirhos, Temme, Rauber, Sch{\"u}tt, Bethge, and
  Wichmann]{geirhos2018generalisation}
Robert Geirhos, Carlos~RM Temme, Jonas Rauber, Heiko~H Sch{\"u}tt, Matthias
  Bethge, and Felix~A Wichmann.
\newblock Generalisation in humans and deep neural networks.
\newblock In \emph{Advances in Neural Information Processing Systems}, pp.\
  7538--7550, 2018.

\bibitem[Geirhos et~al.(2019)Geirhos, Rubisch, Michaelis, Bethge, Wichmann, and
  Brendel]{geirhos2018imagenettrained}
Robert Geirhos, Patricia Rubisch, Claudio Michaelis, Matthias Bethge, Felix~A.
  Wichmann, and Wieland Brendel.
\newblock Imagenet-trained {CNN}s are biased towards texture; increasing shape
  bias improves accuracy and robustness.
\newblock In \emph{International Conference on Learning Representations}, 2019.
\newblock URL \url{https://openreview.net/forum?id=Bygh9j09KX}.

\bibitem[Goodfellow et~al.(2015)Goodfellow, Shlens, and
  Szegedy]{goodfellow_explaining_2015}
Ian~J. Goodfellow, Jonathon Shlens, and Christian Szegedy.
\newblock Explaining and {Harnessing} {Adversarial} {Examples}.
\newblock In \emph{{International} {Conference} on {Learning} {Representations}
  (ICLR)}, 2015.

\bibitem[Hendrycks \& Dietterich(2019)Hendrycks and
  Dietterich]{hendrycks_benchmarking_2018}
Dan Hendrycks and Thomas~G. Dietterich.
\newblock Benchmarking neural network robustness to common corruptions and
  perturbations.
\newblock \emph{International Conference on Learning Representations (ICLR)},
  2019.
\newblock URL \url{http://arxiv.org/abs/1903.12261}.

\bibitem[Hendrycks et~al.(2019)Hendrycks, Mazeika, Kadavath, and
  Song]{hendrycks2019using}
Dan Hendrycks, Mantas Mazeika, Saurav Kadavath, and Dawn Song.
\newblock Using self-supervised learning can improve model robustness and
  uncertainty.
\newblock In \emph{Advances in Neural Information Processing Systems}, pp.\
  15637--15648, 2019.

\bibitem[Hendrycks et~al.(2020)Hendrycks, Basart, Mu, Kadavath, Wang, Dorundo,
  Desai, Zhu, Parajuli, Guo, et~al.]{hendrycks2020many}
Dan Hendrycks, Steven Basart, Norman Mu, Saurav Kadavath, Frank Wang, Evan
  Dorundo, Rahul Desai, Tyler Zhu, Samyak Parajuli, Mike Guo, et~al.
\newblock The many faces of robustness: A critical analysis of
  out-of-distribution generalization.
\newblock \emph{arXiv preprint arXiv:2006.16241}, 2020.

\bibitem[Hendrycks* et~al.(2020)Hendrycks*, Mu*, Cubuk, Zoph, Gilmer, and
  Lakshminarayanan]{hendrycks2019augmix}
Dan Hendrycks*, Norman Mu*, Ekin~Dogus Cubuk, Barret Zoph, Justin Gilmer, and
  Balaji Lakshminarayanan.
\newblock Augmix: A simple method to improve robustness and uncertainty under
  data shift.
\newblock In \emph{International Conference on Learning Representations}, 2020.
\newblock URL \url{https://openreview.net/forum?id=S1gmrxHFvB}.

\bibitem[Hu et~al.(2018)Hu, Shen, and Sun]{hu2018squeeze}
Jie Hu, Li~Shen, and Gang Sun.
\newblock Squeeze-and-excitation networks.
\newblock In \emph{Proceedings of the IEEE conference on computer vision and
  pattern recognition}, pp.\  7132--7141, 2018.

\bibitem[Huang \& Belongie(2017)Huang and Belongie]{huang2017arbitrary}
Xun Huang and Serge Belongie.
\newblock Arbitrary style transfer in real-time with adaptive instance
  normalization.
\newblock In \emph{Proceedings of the IEEE International Conference on Computer
  Vision}, pp.\  1501--1510, 2017.

\bibitem[Ilyas et~al.(2019)Ilyas, Santurkar, Tsipras, Engstrom, Tran, and
  Madry]{ilyas_adversarial_2019}
Andrew Ilyas, Shibani Santurkar, Dimitris Tsipras, Logan Engstrom, Brandon
  Tran, and Aleksander Madry.
\newblock Adversarial examples are not bugs, they are features.
\newblock In \emph{Advances in Neural Information Processing Systems}, pp.\
  125--136, 2019.

\bibitem[Jo \& Bengio(2017)Jo and Bengio]{jo2017measuring}
Jason Jo and Yoshua Bengio.
\newblock Measuring the tendency of cnns to learn surface statistical
  regularities.
\newblock \emph{arXiv preprint arXiv:1711.11561}, 2017.

\bibitem[Kurakin et~al.(2017)Kurakin, Goodfellow, and
  Bengio]{kurakin_adversarial_2016}
Alexey Kurakin, Ian Goodfellow, and Samy Bengio.
\newblock Adversarial examples in the physical world.
\newblock \emph{International Conference on Learning Representations
  (Workshop)}, April 2017.

\bibitem[Lee \& Kolter(2019)Lee and Kolter]{lee_2019}
Mark Lee and J.~Zico Kolter.
\newblock On physical adversarial patches for object detection.
\newblock \emph{International Conference on Machine Learning (Workshop)}, 2019.
\newblock URL \url{http://arxiv.org/abs/1906.11897}.

\bibitem[Li et~al.(2020)Li, Wu, Lim, Belongie, and Weinberger]{li2020feature}
Boyi Li, Felix Wu, Ser-Nam Lim, Serge Belongie, and Kilian~Q. Weinberger.
\newblock On feature normalization and data augmentation, 2020.

\bibitem[Liu et~al.(2017)Liu, Cheng, Hu, Wang, and Bai]{liu2017richer}
Yun Liu, Ming-Ming Cheng, Xiaowei Hu, Kai Wang, and Xiang Bai.
\newblock Richer convolutional features for edge detection.
\newblock In \emph{Proceedings of the IEEE conference on computer vision and
  pattern recognition}, pp.\  3000--3009, 2017.

\bibitem[Lopes et~al.(2019)Lopes, Yin, Poole, Gilmer, and
  Cubuk]{lopes2019improving}
Raphael~Gontijo Lopes, Dong Yin, Ben Poole, Justin Gilmer, and Ekin~D Cubuk.
\newblock Improving robustness without sacrificing accuracy with patch gaussian
  augmentation.
\newblock \emph{arXiv preprint arXiv:1906.02611}, 2019.

\bibitem[Luo et~al.(2019)Luo, Cai, Zhang, Chen, He, and Wang]{luo2019defective}
Tiange Luo, Tianle Cai, Mengxiao Zhang, Siyu Chen, Di~He, and Liwei Wang.
\newblock Defective convolutional layers learn robust cnns.
\newblock \emph{arXiv preprint arXiv:1911.08432}, 2019.

\bibitem[Michaelis et~al.(2019)Michaelis, Mitzkus, Geirhos, Rusak, Bringmann,
  Ecker, Bethge, and Brendel]{michaelis2019benchmarking}
Claudio Michaelis, Benjamin Mitzkus, Robert Geirhos, Evgenia Rusak, Oliver
  Bringmann, Alexander~S Ecker, Matthias Bethge, and Wieland Brendel.
\newblock Benchmarking robustness in object detection: Autonomous driving when
  winter is coming.
\newblock \emph{arXiv preprint arXiv:1907.07484}, 2019.

\bibitem[Nam et~al.(2019)Nam, Lee, Park, Yoon, and Yoo]{nam2019reducing}
Hyeonseob Nam, HyunJae Lee, Jongchan Park, Wonjun Yoon, and Donggeun Yoo.
\newblock Reducing domain gap via style-agnostic networks.
\newblock \emph{arXiv preprint arXiv:1910.11645}, 2019.

\bibitem[Qiao et~al.(2019)Qiao, Wang, Liu, Shen, and Yuille]{qiao2019weight}
Siyuan Qiao, Huiyu Wang, Chenxi Liu, Wei Shen, and Alan Yuille.
\newblock Weight standardization.
\newblock \emph{arXiv preprint arXiv:1903.10520}, 2019.

\bibitem[Rusak et~al.(2020)Rusak, Schott, Zimmermann, Bitterwolf, Bringmann,
  Bethge, and Brendel]{Rusak2020a}
E.~Rusak, L.~Schott, R.~Zimmermann, J.~Bitterwolf, O.~Bringmann, M.~Bethge, and
  W.~Brendel.
\newblock Increasing the robustness of dnns against image corruptions by
  playing the game of noise.
\newblock \emph{arXiv}, Jan 2020.
\newblock URL \url{https://arxiv.org/abs/2001.06057}.

\bibitem[Shi et~al.(2020)Shi, Zhang, Dai, Zhu, Mu, and
  Wang]{shi2020informative}
Baifeng Shi, Dinghuai Zhang, Qi~Dai, Zhanxing Zhu, Yadong Mu, and Jingdong
  Wang.
\newblock Informative dropout for robust representation learning: A shape-bias
  perspective.
\newblock In \emph{International Conference on Machine Learning}, pp.\
  8828--8839. PMLR, 2020.

\bibitem[Szegedy et~al.(2014)Szegedy, Zaremba, Sutskever, Bruna, Erhan,
  Goodfellow, and Fergus]{szegedy_intriguing_2013}
Christian Szegedy, Wojciech Zaremba, Ilya Sutskever, Joan Bruna, Dumitru Erhan,
  Ian Goodfellow, and Rob Fergus.
\newblock Intriguing properties of neural networks.
\newblock In \emph{International Conference on Learning Representations
  (ICLR)}, 2014.

\bibitem[Wang et~al.(2019)Wang, Ge, Lipton, and Xing]{NIPS2019_9237}
Haohan Wang, Songwei Ge, Zachary Lipton, and Eric~P Xing.
\newblock Learning robust global representations by penalizing local predictive
  power.
\newblock In \emph{Advances in Neural Information Processing Systems}, pp.\
  10506--10518, 2019.

\bibitem[Wu \& He(2018)Wu and He]{wu2018group}
Yuxin Wu and Kaiming He.
\newblock Group normalization.
\newblock In \emph{Proceedings of the European Conference on Computer Vision
  (ECCV)}, pp.\  3--19, 2018.

\bibitem[Xie et~al.(2019)Xie, Hovy, Luong, and Le]{xie2019self}
Qizhe Xie, Eduard Hovy, Minh-Thang Luong, and Quoc~V Le.
\newblock Self-training with noisy student improves imagenet classification.
\newblock \emph{arXiv preprint arXiv:1911.04252}, 2019.

\bibitem[Xie \& Tu(2015)Xie and Tu]{xie2015holistically}
Saining Xie and Zhuowen Tu.
\newblock Holistically-nested edge detection.
\newblock In \emph{Proceedings of the IEEE international conference on computer
  vision}, pp.\  1395--1403, 2015.

\bibitem[Yin et~al.(2019)Yin, Lopes, Shlens, Cubuk, and Gilmer]{yin2019fourier}
Dong Yin, Raphael~Gontijo Lopes, Jon Shlens, Ekin~Dogus Cubuk, and Justin
  Gilmer.
\newblock A fourier perspective on model robustness in computer vision.
\newblock In \emph{Advances in Neural Information Processing Systems}, pp.\
  13255--13265, 2019.

\bibitem[Yun et~al.(2019)Yun, Han, Oh, Chun, Choe, and Yoo]{yun2019cutmix}
Sangdoo Yun, Dongyoon Han, Seong~Joon Oh, Sanghyuk Chun, Junsuk Choe, and
  Youngjoon Yoo.
\newblock Cutmix: Regularization strategy to train strong classifiers with
  localizable features.
\newblock In \emph{Proceedings of the IEEE International Conference on Computer
  Vision}, pp.\  6023--6032, 2019.

\bibitem[Zhang(2019)]{zhang2019shiftinvar}
Richard Zhang.
\newblock Making convolutional networks shift-invariant again.
\newblock In \emph{ICML}, 2019.

\end{thebibliography}
\bibliographystyle{iclr2021_conference}

\newpage
\clearpage
\renewcommand{\thetable}{A\arabic{table}}
\renewcommand{\thefigure}{A\arabic{figure}}
\setcounter{table}{0}
\setcounter{figure}{0}
\setcounter{section}{0}


\appendix
\section{Appendix}

\subsection{Details of the edge detector}
\label{app_sec:edge_dataset}
\begin{figure}[h]
	\begin{center}
		\includegraphics[width=0.9\linewidth, height=0.16\textheight]{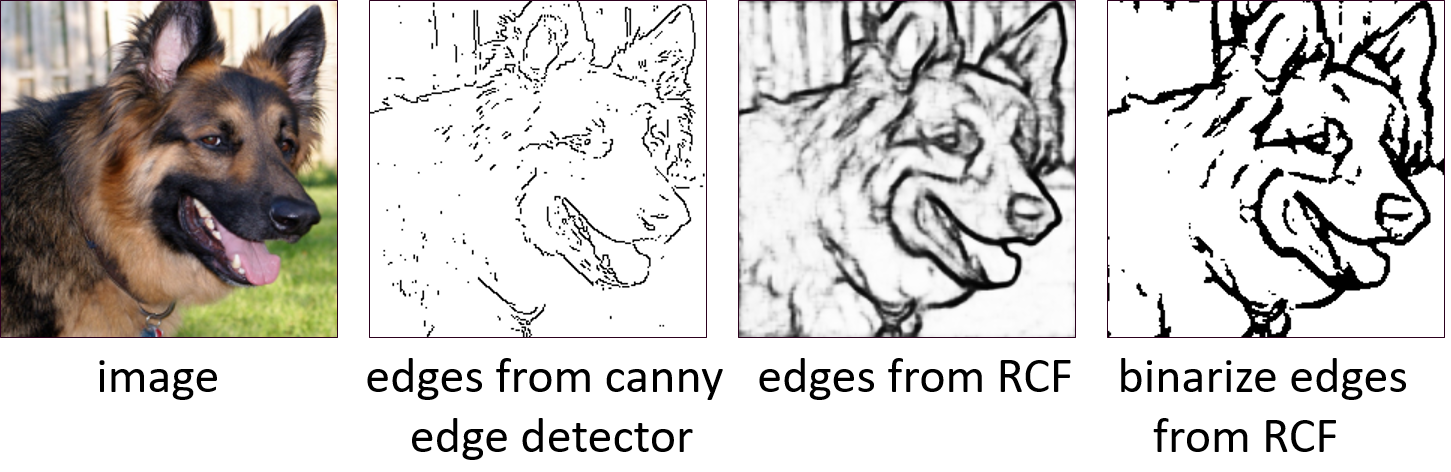}
		\caption{Illustration of a natural image, and edge maps extracted with Canny edge detector \cite{canny1986computational}, with RCF \cite{liu2017richer}, and with binarized edges from RCF.}
		\vskip -0.15in
		\label{fig:edgemaps}
	\end{center}
\end{figure}

The edge details are extracted by the CNN-based edge detector using richer convolutional features (RCF) proposed in \cite{liu2017richer}, which produces edge maps as side outputs of different layers of the network. 
The level of details of the edge maps gradually become coarser from the early layers to the final layers of the network. 
We employ here a ResNet101-based RCF edge detector which was trained on the BSDS500 dataset \cite{arbelaez2010contour} and use the pre-trained network to extract edge maps from images.
The ResNet-101 architecture is composed of four stages, and we select the side output of the second stage to generate the edge maps, as we observed that these edge maps contain the essential shape details, as shown in Figure \ref{fig:edgemaps}.

\subsection{ImageNet20 dataset}
\label{app_sec:IN_dataset}
We use a subset of \(20\) classes from ImageNet dataset to study the influence of the shape bias on robustness towards corruptions. It comprises a total of 25784 training and 1000 validation images. This subset consists of animal classes (\texttt{african elephant}, \texttt{german shepherd}, \texttt{tabby cat}, \texttt{arabian camel}, \texttt{tailed frog}, \texttt{scorpion}), birds (\texttt{king penguin}, \texttt{albatross}), insects (\texttt{fly}, \texttt{sulphur butterfly}), man made objects (\texttt{tea pot}, \texttt{stop watch}, \texttt{teddy bear}, \texttt{fur coat}), automobile (\texttt{sports car}, \texttt{trolley bus}, \texttt{life boat}) and edible items (\texttt{mushroom}, \texttt{bell pepper}, \texttt{pretzel}).
\subsection{Techniques for the evaluation of shape bias}
\begin{figure}[tb]
\centering
\begin{subfigure}[h]{\linewidth}
\includegraphics[width=\linewidth, height=0.14\textheight]{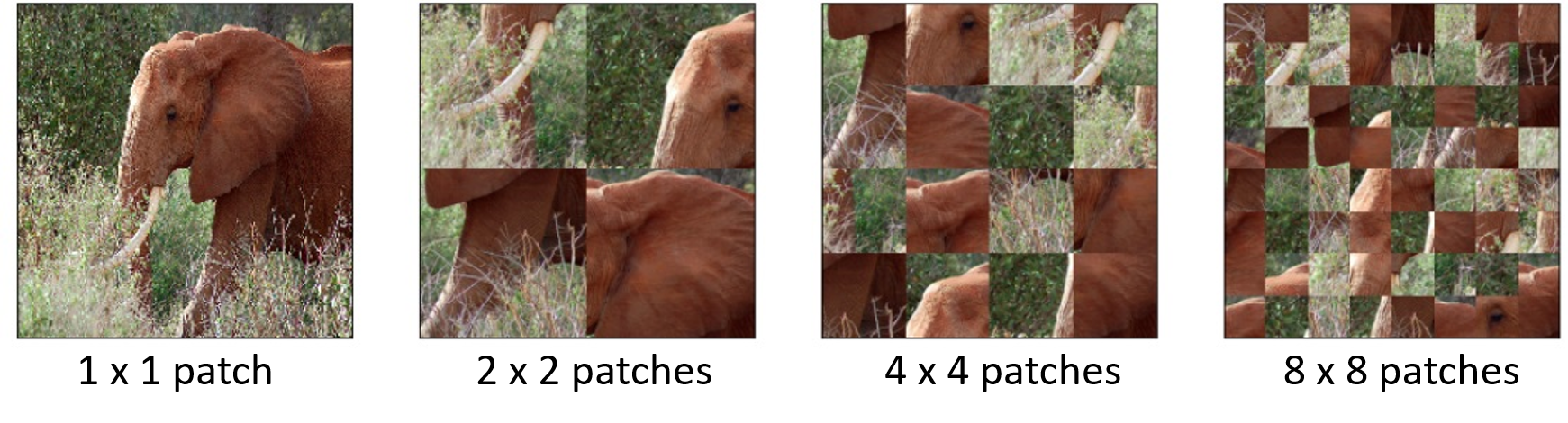}
\caption{The image is divided into $n \times n$ patches and the patches are randomly shuffled. The global object shape is increasingly perturbed with larger n.}\label{fig:shuffled_patches}
\end{subfigure}

\begin{subfigure}[b]{\linewidth}
\includegraphics[width=0.8\linewidth, height=0.14\textheight]{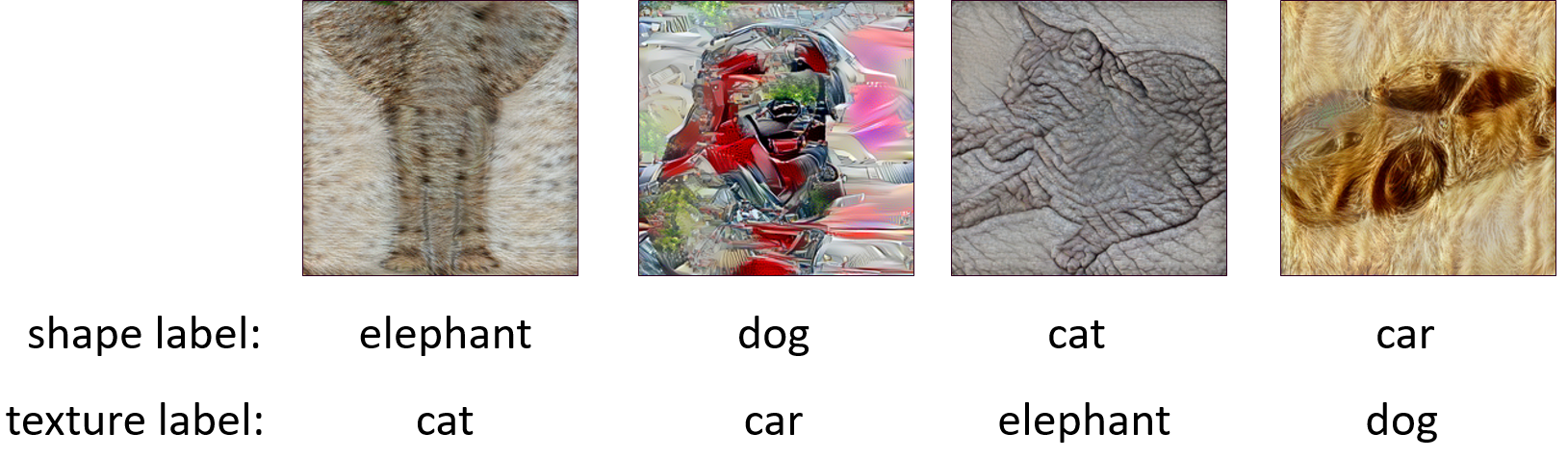}
\caption{Images with conflicting shape and texture cues. The images are obtained by applying style transfer with a texture image as style source.}\label{fig:cue_conflict}
\end{subfigure}
\caption{Techniques to evaluate the shape bias of networks}
\label{fig:eval_shape_bias}
\end{figure}

Figure \ref{fig:eval_shape_bias} depicts the two techniques that are used in our work to evaluate the shape bias. Figure \ref{fig:shuffled_patches} points to the \textbf{shuffled image patches} technique that perturb the shape details by preserving the local texture. On the other hand, Figure \ref{fig:cue_conflict} show examples of \textbf{texture-shape cue conflict images} that test the network's bias towards shape or texture.

\subsection{Convolutional filters}

\begin{figure}
\centering
\begin{subfigure}[h]{.3\linewidth}
\includegraphics[width=\linewidth]{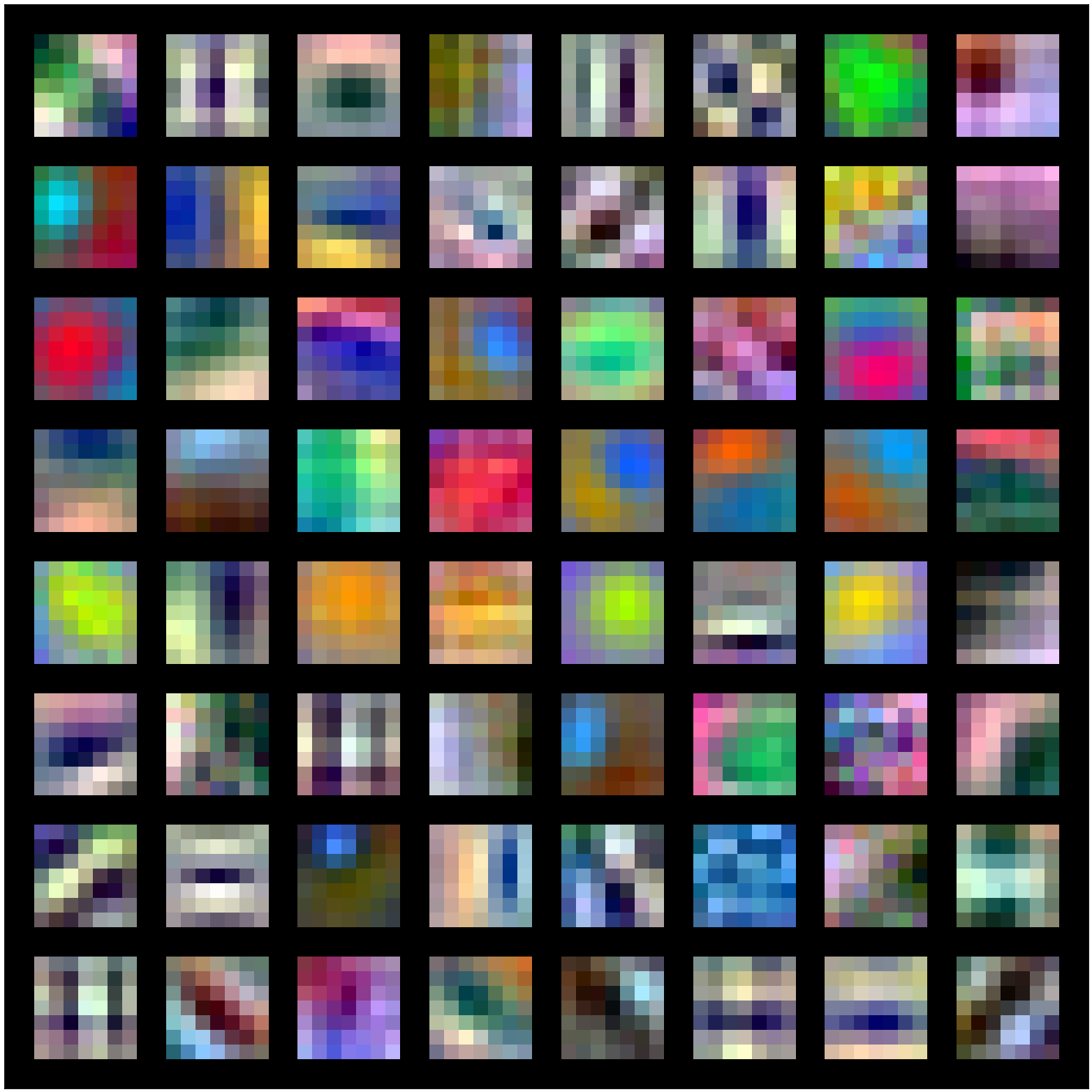}
\caption{Standard CNN (IN)}\label{fig:conv1_filter_standard}
\end{subfigure}
\begin{subfigure}[h]{.3\linewidth}
\includegraphics[width=\linewidth]{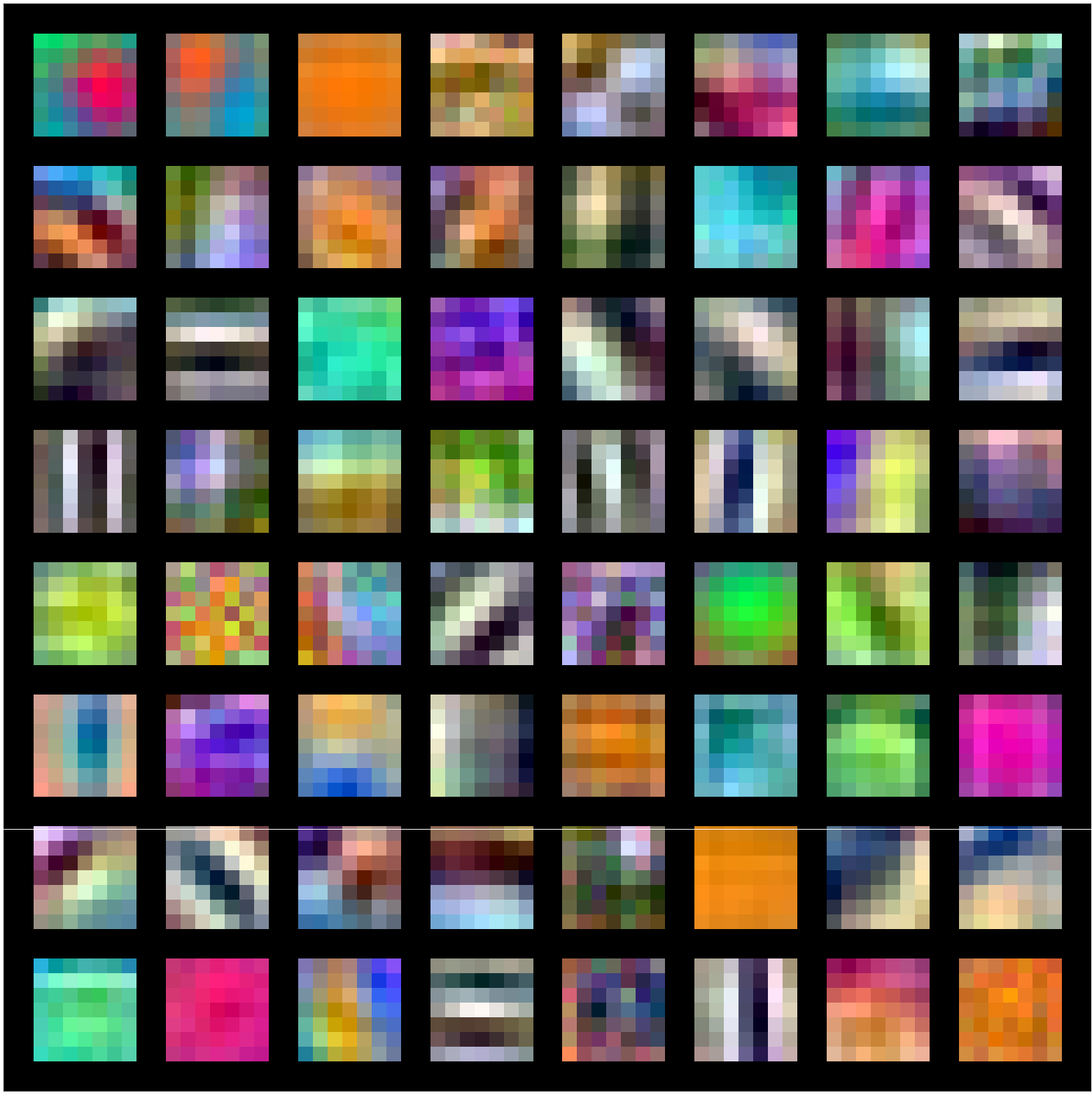}
\caption{Stylized CNN (SIN)}\label{fig:conv1_filter_stylized}
\end{subfigure}
\begin{subfigure}[h]{.3\linewidth}
\includegraphics[width=\linewidth]{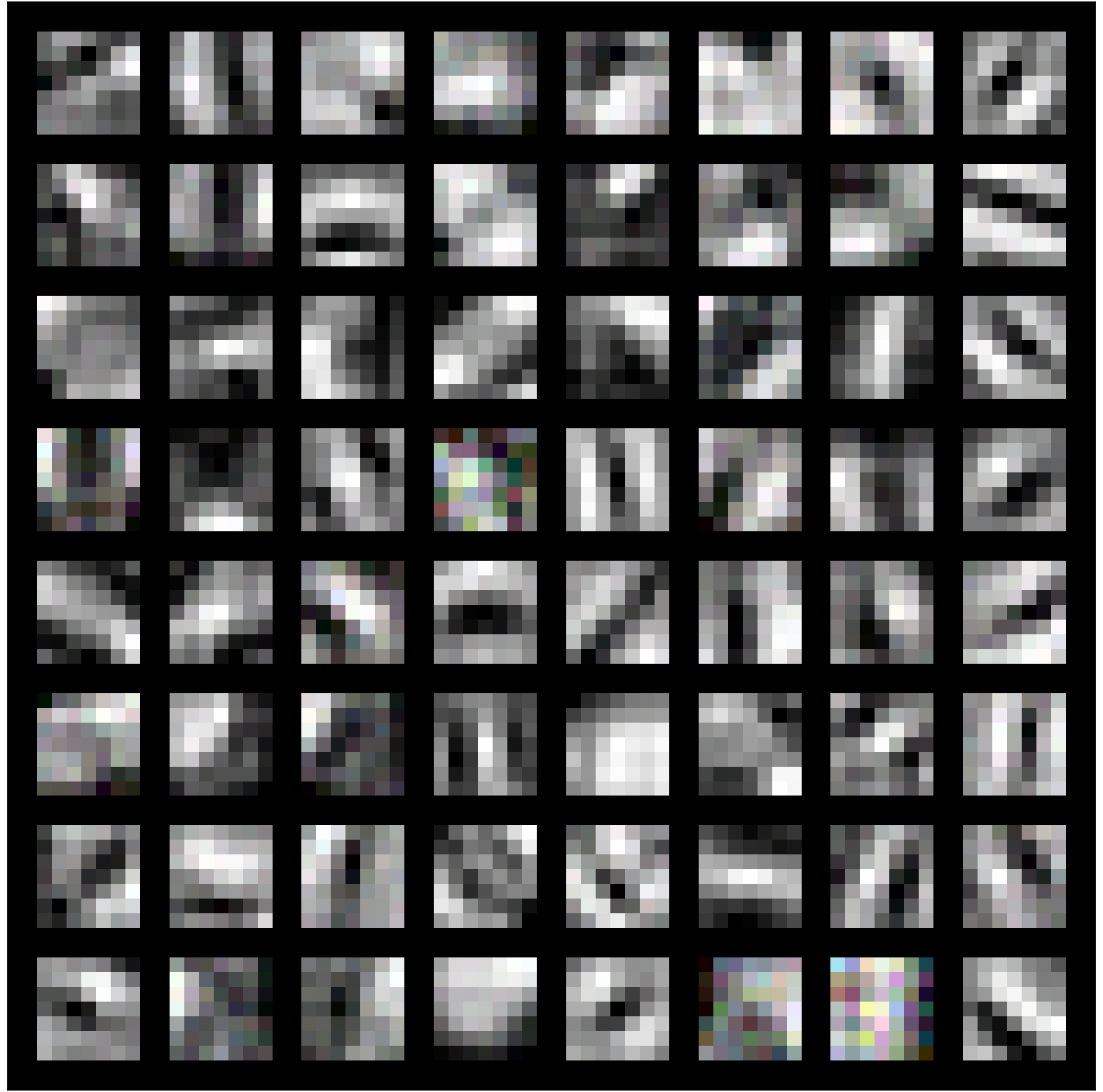}
\caption{Edge CNN (E)}\label{fig:conv1_filter_edge}
\end{subfigure}
\caption{Filters of first convolutional layer of Standard CNN (IN), Stylized CNN (SIN), Edge CNN (E) (ours). The filters of E strongly resembles Gabor filters for edge detection than other two networks.}
\label{fig:conv1_filter}
\end{figure}

We show in Table \ref{tab:stylization} and Table \ref{tab:shape_bias} that the network E demonstrates stronger shape bias than IN and SIN. In Figure \ref{fig:conv1_filter}, we visualize the filters of first convolutional layer of networks IN, SIN and E to understand the behavior of these networks. As seen in the figure, filters of E strongly resembles the Gabor filters for edge detection compared to other networks. These results suggest that E extracts features that corresponds to the shape information in the form of edges and effectively based its decision on shape details. The filters of E are non-colored because E is trained on edge maps whereas IN and SIN are trained on natural and stylized images respectively.

\subsection{Additional experimental results}
\label{app_sec:extended experiments}
\subsubsection{Patch shuffled results on the whole validation data}
We extend the results of \emph{shuffled image patches} in Table \ref{tab:stylization} and \ref{tab:shape_bias} on the whole validation data and are shown in Table \ref{tab_app:stylization} and \ref{tab_app:shape_bias} respectively.

\begin{table}[t]
\centering
\begin{tabular}{|l||l|l|l||}
\hline
 & \multicolumn{3}{c||}{\thead{shuffled image patches $4 \times 4$ (\%)}} \\  \cline{2-4} 
\thead{Network\\} &   \thead{No styling} & \thead{style blending} & \thead{style randomization} \\ \hline
\makecell{IN}   & \makecell{57.8}  & \makecell{44.9}  & \makecell{36} 
\\ \hline
\makecell{SIN} & \makecell{33.1}  & \makecell{32.2}   & \makecell{29.8} 
\\ \hline
\makecell{E}  & \makecell{28.3}    & \makecell{28.4}   & \makecell{\bf 23.43}
\\ \hline
\end{tabular}
\caption{Comparison of different feature space style augmentation methods on $4 \times 4$ shuffled image patches on whole validation data.}
	\vskip -0.15in
	\label{tab_app:stylization}
\end{table}

\begin{table}[t]
\centering
\begin{tabular}{|l||l|l|l||}
\hline
 & \multicolumn{3}{c||}{\thead{shuffled image patches (\%)}}  \\  \cline{2-4} 
\thead{Network\\} &   \thead{\textbf{$2 \times 2$}}  & \thead{$4 \times 4$}  & \thead{$8 \times 8$}\\ \hline
\makecell{IN}   & \makecell{66.7}  & \makecell{36}  & \makecell{26.7} 
\\ \hline
\makecell{SIN} & \makecell{63.5}  & \makecell{29.8}   & \makecell{17.6} 
\\ \hline
\makecell{E}  & \makecell{61.8}    & \makecell{23.4}   & \makecell{9.8} 
\\ \hline
\makecell{SE}  & \makecell{\bf 57}  & \makecell{23.7} & \makecell{10}  
\\ \hline
\makecell{E-SIN} & \makecell{59} & \makecell{\bf 20.1} & \makecell{\bf 9.1} 
\\ \hline

\end{tabular}
\caption{Comparison of models trained on different datasets on shuffled image patches evaluated on whole validation data.}
	\label{tab_app:shape_bias}
\end{table}

\subsubsection{ResNet18 on ImageNet200}
We present an additional evaluation on ImageNet200 (200 TinyImageNet classes but all the images in full resolution from ImageNet) in Table \ref{app_tab:IN200}: E-SIN has higher shape bias but demonstrates lower mean Corruption Accuracy(mCA) than SIN. This reinforces our core finding of shape bias not implying corruption robustness on a larger dataset. Here, all models have clean validation accuracy of about 70\%.

\begin{table}[t]
\centering
\begin{tabular}{|l||l|l|l||l|l|}
\hline
 & \multicolumn{3}{c||}{\thead{shuffled image patches(\%)}}  &    & \\  \cline{2-4} 
\thead{Network\\} &   \thead{\textbf{$2 \times 2$}}  & \thead{$4 \times 4$}  & \thead{$8 \times 8$} & \thead{Cue conflict \\ shape \#880} & \thead{Mean corruption \\ accuracy (\%)}\\ \hline
\makecell{IN}   & \makecell{78.6}  & \makecell{45.7}  & \makecell{15.6} & \makecell{90}  & \makecell{35.8}  
\\ \hline
\makecell{SIN} & \makecell{61.7}  & \makecell{17.7}   & \makecell{3.7} & \makecell{273}     & \makecell{\bf 52.4}
\\ \hline
\makecell{E-SIN}  & \makecell{\bf 54.3}  & \makecell{\bf 10.6} & \makecell{\bf 1.8}  & \makecell{\bf 337}     &  \makecell{47.7} 
\\ \hline
\end{tabular}
\caption{Shape based evaluation \& corruption accuracies with ResNet18 on 200 classes of ImageNet. Patch shuffled evaluation is conducted on 5474 correctly classified validation images by all the networks.}
	\label{app_tab:IN200}
\end{table}

\subsubsection{ResNet18 with BatchNorm on ImageNet20}
We provide results on ResNet18 wih BatchNorm in Table \ref{app_tab:resnet18_BN} showing that it leads to the same findings as the ones reported in the paper for Group Normalization + Weight Standardization: Both networks E and E-SIN exhibit more shape bias but lower mean Corruption Accuracy (mCA) than SIN. Superposition (SE+IN) has comparable mean corruption accuracy with SIN despite showing lower shape bias on patch shuffled and cue conflict evaluation. Here, all models below exhibit validation accuracies of 89\%-90\%. Note that the architecture ResNet18 wih BatchNorm does not include Group Normalization and Weight Standardization layers.

\begin{table}[t]
\centering
\begin{tabular}{|l||l|l|l||l|l|}
\hline
 & \multicolumn{3}{c||}{\thead{shuffled image patches(\%)}}  &    & \\  \cline{2-4} 
\thead{Network\\} &   \thead{\textbf{$2 \times 2$}}  & \thead{$4 \times 4$}  & \thead{$8 \times 8$} & \thead{Cue conflict \\ shape \#400} & \thead{Mean corruption \\ accuracy (\%)}\\ \hline
\makecell{IN}   & \makecell{89.9}  & \makecell{72.9}  & \makecell{41.9} & \makecell{58}  & \makecell{57.8}  
\\ \hline
\makecell{SIN} & \makecell{\bf 81.3}  & \makecell{47.7}   & \makecell{19.6} & \makecell{151}     & \makecell{\bf 72.9}
\\ \hline
\makecell{E}  & \makecell{\bf 81.3}    & \makecell{\bf 39}   & \makecell{\bf 9.5} & \makecell{128}    & \makecell{50.8} 
\\ \hline
\makecell{E-SIN}  & \makecell{84.3}  & \makecell{41} & \makecell{11.2}  & \makecell{\bf 169}     &  \makecell{67.5} 
\\ \hline
\makecell{SE+IN} & \makecell{86.1} & \makecell{60.8} & \makecell{33.5} & \makecell{91}   & \makecell{\bf 72} 
\\ \hline

\end{tabular}
\caption{Shape based evaluation \& corruption accuracies of ResNet18 networks with BatchNorm. Patch shuffled evaluation is conducted on 775 correctly classified validation images by all the networks.}
	\label{app_tab:resnet18_BN}
\end{table}


\subsubsection{ResNet50 on ImageNet20}
We present our analysis on deeper architecuture like ResNet50 with Group Normalization + Weight Standardization on ImageNet20 in Table \ref{app_tab:resnet50}: E and E-SIN with higher shape bias have lower mean corruption accuracy than SIN, whereas SE+IN with lower shape bias reach similar accuracy as SIN. These results show that our findings apply to deeper architectures. Here, all models have 86\%-87\% clean validation accuracy.

\begin{table}[t]
\centering
\begin{tabular}{|l||l|l|l||l|l|}
\hline
 & \multicolumn{3}{c||}{\thead{shuffled image patches(\%)}}  &    & \\  \cline{2-4} 
\thead{Network\\} &   \thead{\textbf{$2 \times 2$}}  & \thead{$4 \times 4$}  & \thead{$8 \times 8$} & \thead{Cue conflict \\ shape \#400} & \thead{Mean corruption \\ accuracy (\%)}\\ \hline
\makecell{IN}   & \makecell{88.8}  & \makecell{74.1}  & \makecell{50.7} & \makecell{65}  & \makecell{58}  
\\ \hline
\makecell{SIN} & \makecell{83.2}  & \makecell{46}   & \makecell{20.5} & \makecell{141}     & \makecell{\bf 78.2}
\\ \hline
\makecell{E}  & \makecell{\bf 72.5}    & \makecell{34.5}   & \makecell{13.7} & \makecell{164}    & \makecell{52.2} 
\\ \hline
\makecell{E-SIN}  & \makecell{73.8}  & \makecell{\bf 25.7} & \makecell{\bf 8.4}  & \makecell{\bf 209}     &  \makecell{72.3} 
\\ \hline
\makecell{SE+IN} & \makecell{85.5} & \makecell{66.4} & \makecell{36.5} & \makecell{102}   & \makecell{\bf 78.1} 
\\ \hline

\end{tabular}
\caption{Shape based evaluation \& corruption accuracies with ResNet50 architecture.}
	\label{app_tab:resnet50}
\end{table}


\subsubsection{DenseNet121 and MobileNetV2 on ImageNet20}
We extend our findings from ResNet to other architectures like DenseNet121 and MobileNetV2 on ImageNet20 with style randomization. Table \ref{app_tab:densenet121} and Table \ref{app_tab:mobilenetv2} present the shape-based evaluation and mean corruption accuracy (mCA) of DenseNet121 and MobileNetV2 on different settings, respectively. Similar to the results on ResNet, we find that E-SIN exhibits a higher shape bias than other settings but still has a lower mCA than SIN. On the other hand, Superposition (SE+IN) shows a lower shape bias with similar mCA to SIN. These results show that our conclusion is also valid using different neural architectures.

Note that these results also include additional intermediate settings E-IN and E+IN. The explanation of these two settings is as follows:
\begin{itemize}[noitemsep,topsep=0pt,parsep=0pt,partopsep=0pt, leftmargin=*]
	\item E-IN: This is a network training setting similar to E-SIN. In E-SIN, network is pretrained on Edge dataset (E) in the first stage of training for 75 epochs and later finetune on both the stylized images (SIN) and original images (IN) for another 75 epochs in the second stage (please refer Section \ref{sec:experiments} for training details). Similarly, E-IN also pretrains the network on Edge dataset (E) in the first stage but later finetune only on the original images (IN) in the second stage. This setting also shares similarity with network setting E, where network pretrains on edge maps during the first stage and later finetune on both edge maps and original images in the second stage of training.
	
	\item E+IN: Similar to Superposition (SE+IN), this dataset variant E+IN interpolates edge maps \(I_{\text{E}}\) from Edge dataset (E) with images \(I_{\text{IN}}\) from ImageNet20 (IN): $I_{\text{E+IN}} := (1-\alpha)\cdot I_{\text{E}} + \alpha\cdot I_{\text{IN}}$. We set $\alpha=0.5$. Similar to SE+IN, this setting also pretrains the network on E+IN images in the first stage of training and later finetunes on both E+IN and also on the original images (IN) in the second stage.
	
\end{itemize}

\begin{table}[t]
\centering
\begin{tabular}{|l||l|l|l||l|l|}
\hline
 & \multicolumn{3}{c||}{\thead{shuffled image patches(\%)}}  &    & \\  \cline{2-4} 
\thead{Network\\} &   \thead{\textbf{$2 \times 2$}}  & \thead{$4 \times 4$}  & \thead{$8 \times 8$} & \thead{Cue conflict \\ shape \#400} & \thead{Mean corruption \\ accuracy (\%)}\\ \hline
\makecell{IN}   & \makecell{\bf 62.9}  & \makecell{37.9}  & \makecell{23.5} & \makecell{55}  & \makecell{62.7}  
\\ \hline
\makecell{SIN} & \makecell{68.4}  & \makecell{42.5}   & \makecell{21.7} & \makecell{154}     & \makecell{\bf 80.3}
\\ \hline
\makecell{E}  & \makecell{65.2}    & \makecell{35.8}   & \makecell{14.0} & \makecell{167}    & \makecell{60.2} 
\\ \hline
\makecell{E-SIN}  & \makecell{66.0}  & \makecell{35.4} & \makecell{\bf 12.0}  & \makecell{\bf 219}     &  \makecell{77.7} 
\\ \hline
\makecell{SE+IN} & \makecell{68.3} & \makecell{51.7} & \makecell{32.0} & \makecell{118}   & \makecell{79.0} 
\\ \hline
\makecell{E-IN} & \makecell{64.5} & \makecell{\bf 33.7} & \makecell{12.7} & \makecell{164}   & \makecell{62.4} 
\\ \hline
\makecell{E+IN} & \makecell{69.0} & \makecell{49.4} & \makecell{31.1} & \makecell{69}   & \makecell{65.0} 
\\ \hline

\end{tabular}
\caption{Shape based evaluation \& corruption accuracies with DenseNet121 architecture. All models exhibit clean validation accuracies of 88\%-91\%. Patch shuffled evaluation is conducted on 761 correctly classified validation images by all the networks. It can be observed from these results that the SE+IN have similar mCA to SIN despite having lower shape bias than both SIN and E-SIN.}
	\label{app_tab:densenet121}
\end{table}

\begin{table}[t]
\centering
\begin{tabular}{|l||l|l|l||l|l|}
\hline
 & \multicolumn{3}{c||}{\thead{shuffled image patches(\%)}}  &    & \\  \cline{2-4} 
\thead{Network\\} &   \thead{\textbf{$2 \times 2$}}  & \thead{$4 \times 4$}  & \thead{$8 \times 8$} & \thead{Cue conflict \\ shape \#400} & \thead{Mean corruption \\ accuracy (\%)}\\ \hline
\makecell{IN}   & \makecell{53.2}  & \makecell{27.5}  & \makecell{21.8} & \makecell{90}  & \makecell{56.5}  
\\ \hline
\makecell{SIN} & \makecell{50.6}  & \makecell{20.4}   & \makecell{9.0} & \makecell{189}     & \makecell{\bf 72.2}
\\ \hline
\makecell{E}  & \makecell{52.6}    & \makecell{18.7}   & \makecell{11.4} & \makecell{160}    & \makecell{56.0} 
\\ \hline
\makecell{E-SIN}  & \makecell{\bf 49.1}  & \makecell{\bf 16.1} & \makecell{\bf 6.0}  & \makecell{\bf 212}     &  \makecell{69.0} 
\\ \hline
\makecell{SE+IN} & \makecell{57.4} & \makecell{31.6} & \makecell{20.0} & \makecell{138}   & \makecell{70.0} 
\\ \hline
\makecell{E-IN} & \makecell{52.4} & \makecell{19.0} & \makecell{9.0} & \makecell{148}   & \makecell{56.6} 
\\ \hline
\makecell{E+IN} & \makecell{56.4} & \makecell{33.0} & \makecell{20.0} & \makecell{94}   & \makecell{60.3} 
\\ \hline

\end{tabular}
\caption{Shape based evaluation \& corruption accuracies with MobileNetV2 architecture. All models exhibit clean validation accuracies of 86\%-88.5\%. Patch shuffled evaluation is conducted on 700 correctly classified validation images by all the networks. Here, SE+IN shows high corruption robustness close to the one of SIN despite having a lower shape bias than SIN and E-SIN. The reason for the gap between SIN and SE+IN corruption accuracy can be explained as follows: stylized dataset SIN is seen as strongly augmented dataset than the superposition SE+IN (notice perceptual differences in Figure 2) and unlike larger architectures like ResNet18/DenseNet121, the compact nature of the MobileNetV2 architecture is shown to benefit from such stronger data augmentation than the superposition.}
	\label{app_tab:mobilenetv2}
\end{table}

\subsubsection{Evaluation on edge maps of validation set}
We evaluate the performance of models trained under different training settings on edge map based validation set of ImageNet20 and the validation accuracy results are presented in Table \ref{tab_app:eval_edge_validation}.
Note that among different settings in our experiments, edge maps of training data are directly used in E throughout the training, also used for pretraining the network in E-SIN, and stylized edge maps of training data are used during training of Superposition (SE+IN) (please refer Section A.3 for training details). On the other hand, edge maps are not used in any way for training in IN and SIN. From the results of Table \ref{tab_app:eval_edge_validation}, we can observe that the setting E has the highest edge map-based validation accuracy among all others as the edge maps have been used in the entire training process.

\begin{table}[t]
\centering
\begin{tabular}{|l||l|l|l||}
\hline
 & \multicolumn{3}{c||}{\thead{Architectures}}  \\  \cline{2-4} 
\thead{Network\\} &   \thead{\textbf{ResNet18 (\%)}}  & \thead{DenseNet121 (\%)}  & \thead{MobileNetV2 (\%)}\\ \hline
\makecell{IN}   & \makecell{18.5}  & \makecell{22.1}  & \makecell{31.8} 
\\ \hline
\makecell{SIN} & \makecell{48.6}  & \makecell{51.2}   & \makecell{57.8} 
\\ \hline
\makecell{E}  & \makecell{\bf 77.9}    & \makecell{\bf 80.1}   & \makecell{\bf 74.5} 
\\ \hline
\makecell{E-SIN} & \makecell{72.4} & \makecell{71.9} & \makecell{67.2} 
\\ \hline
\makecell{SE+IN}  & \makecell{47.8}  & \makecell{51.6} & \makecell{54.8}  
\\ \hline
\end{tabular}
\caption{Evaluation of different networks on edge map based validation dataset. It can observed that the network E has the highest edge map-based validation accuracy as the edge maps have been used in the entire training process. Here, SE+IN has higher validation accuracy than IN, comparable to SIN. The reason SE+IN has higher validation accuracy is that the stylized edges are used during training of the SE+IN setting.}
	\label{tab_app:eval_edge_validation}
\end{table}

\subsection{Significance of interpolation parameter in superposition}
\begin{figure}[t]
\centering
\includegraphics[width=0.8\linewidth]{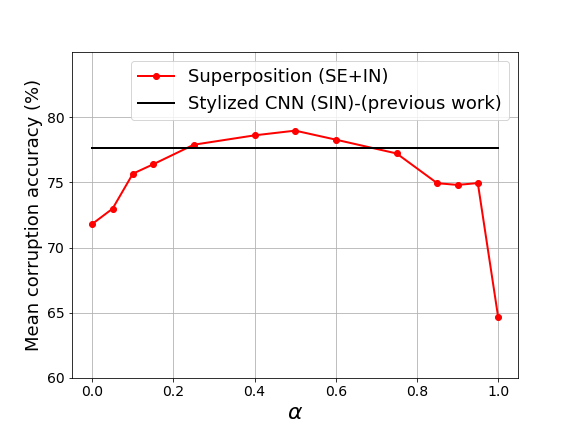}
\caption{Mean corruption accuracy on ImageNet-C corruptions at different values of $\alpha$ in a stylized dataset SE+IN. Here $\alpha \in \{0,\, 0.05,\, 0.1,\, 0.15,\, 0.25,\, 0.4,\, 0.5,\, 0.6,\, 0.75,\, 0.85,\, 0.9,\, 0.95,\, 1\}$. The solid black line represents the performance of the baseline SIN.
}
\label{fig:superposition_alpha}
\end{figure}

In Section \ref{sec:experiments} under \emph{Stylization variants}, we discussed about studying the \textbf{role of natural image statistics} by interpolating images \(I_{\text{SE}}\) from SE with images \(I_{\text{IN}}\) from IN: $I_{\text{SE+IN}} := (1-\alpha)\cdot I_{\text{SE}} + \alpha\cdot I_{\text{IN}}$. We show in Table \ref{tab:finetune_corruptions} that such setup (SE+IN) with $\alpha=0.5$ outperforms SIN despite having lower shape bias than all the other networks. In Figure \ref{fig:superposition_alpha}, we show the performance of SE+IN at different values of $\alpha$ on the mean corruption accuracy on all 15 ImageNet-C corruptions at different severity levels. Here, $\alpha=0$ corresponds to the network trained only on SE whereas $\alpha=1$ on IN. As shown in the figure, the two extreme set of values of $\alpha$ i.e., $\alpha$ being very smaller or very larger result drop in performance on corruptions. This implies that the images with balanced details of natural image statistics and style variations is essential for improved performance on corruptions.

\subsection{Performance of different networks on common corruptions}
\begin{figure}
\centering
\begin{subfigure}[h]{1.0\linewidth}
\includegraphics[width=\linewidth, height=0.19\textheight]{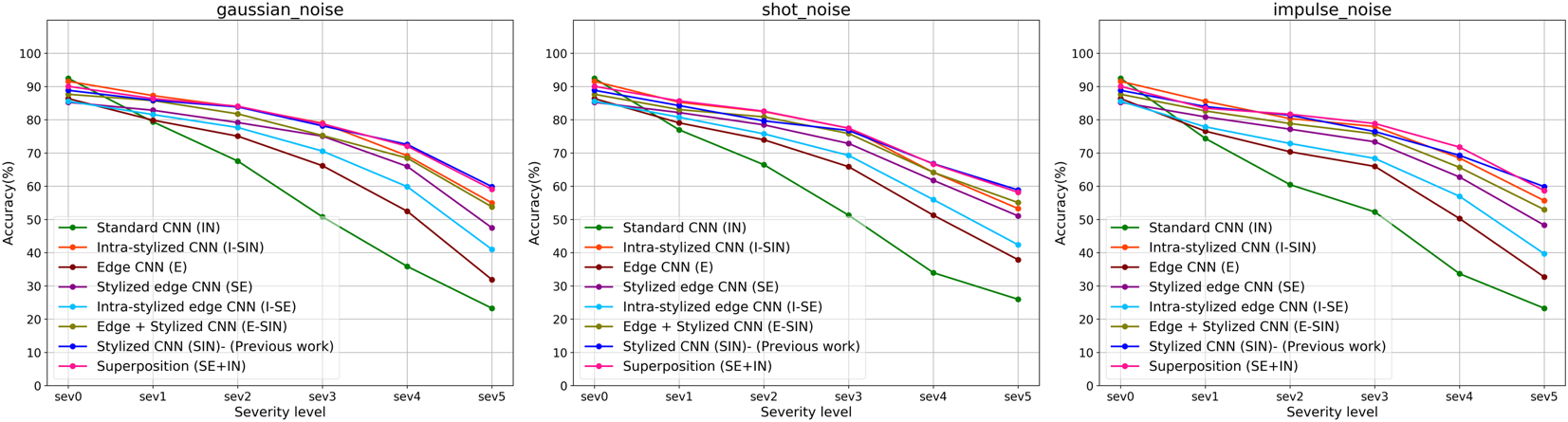}
\end{subfigure}

\begin{subfigure}[h]{1.0\linewidth}
\includegraphics[width=\linewidth, height=0.19\textheight]{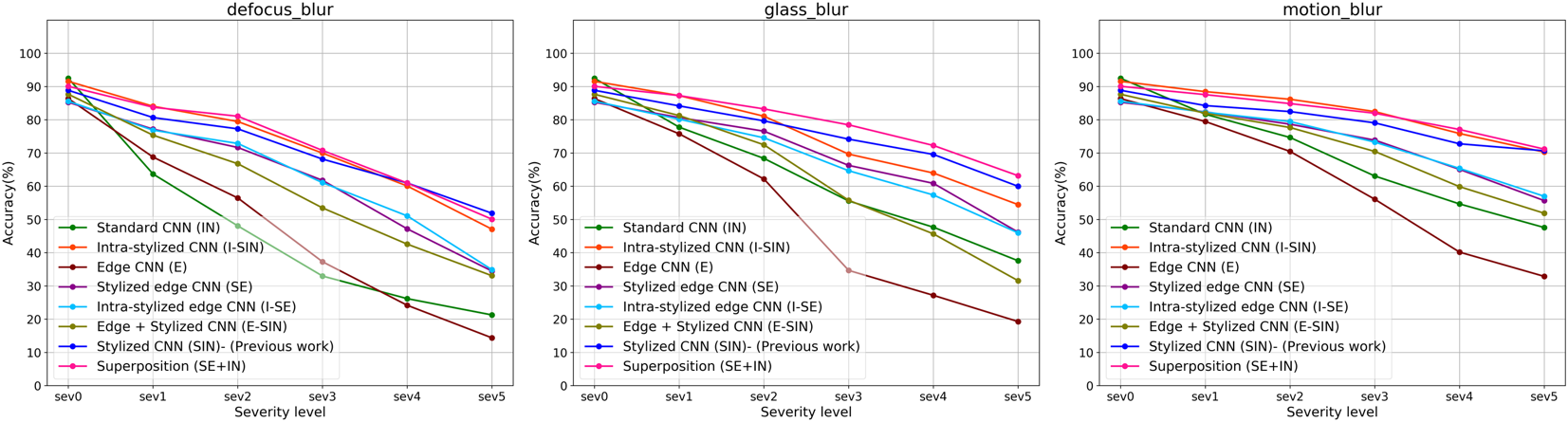}
\end{subfigure}

\begin{subfigure}[h]{1.0\linewidth}
\includegraphics[width=\linewidth, height=0.19\textheight]{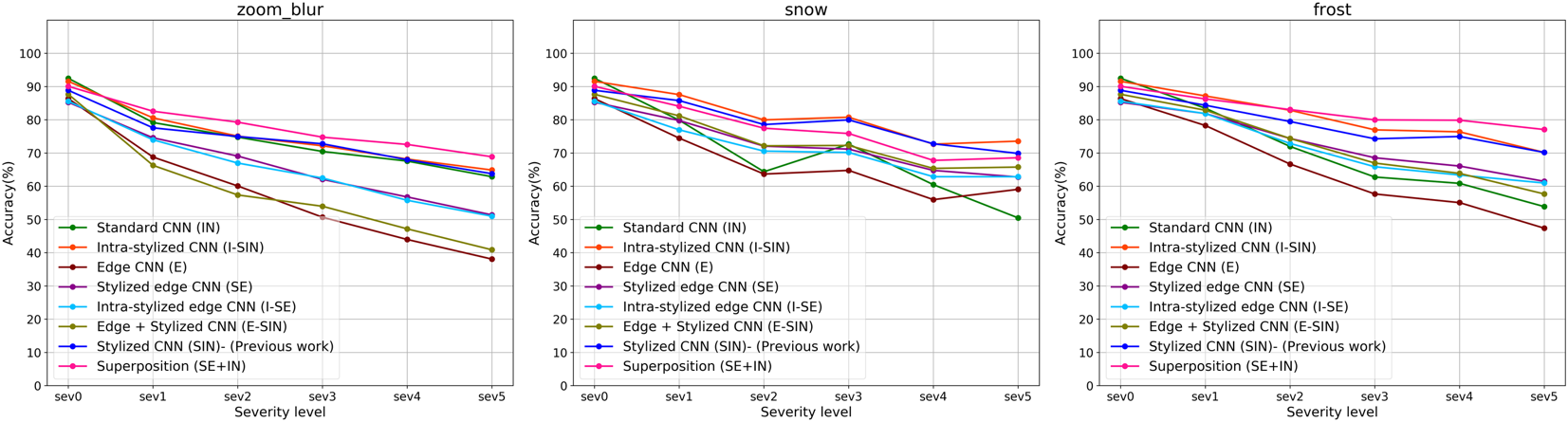}
\end{subfigure}

\begin{subfigure}[h]{1.0\linewidth}
\includegraphics[width=\linewidth, height=0.19\textheight]{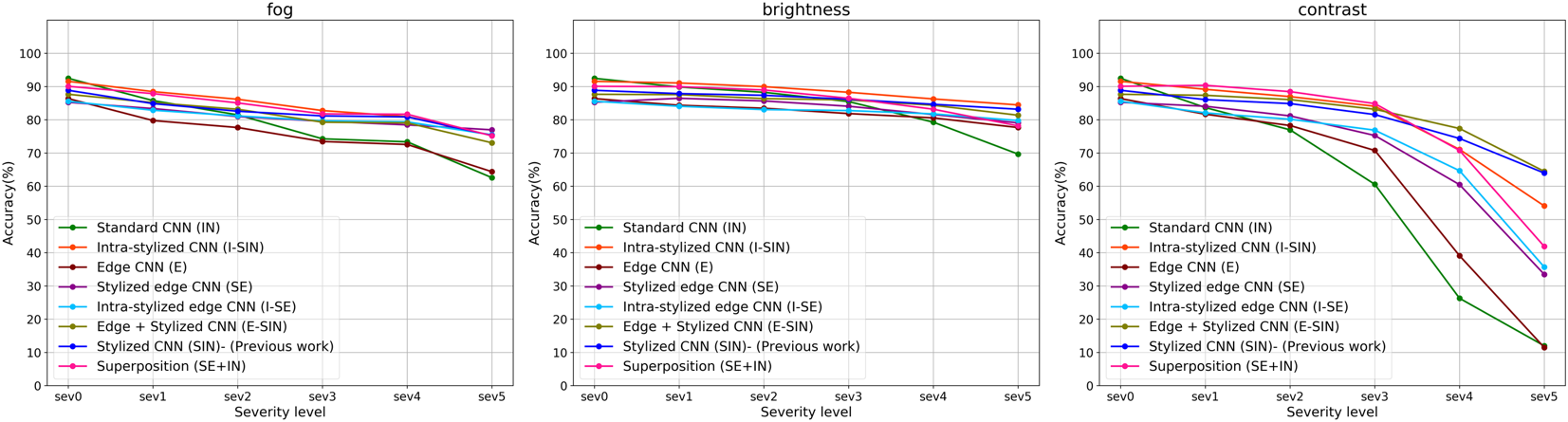}
\end{subfigure}

\begin{subfigure}[h]{1.0\linewidth}
\includegraphics[width=\linewidth, height=0.19\textheight]{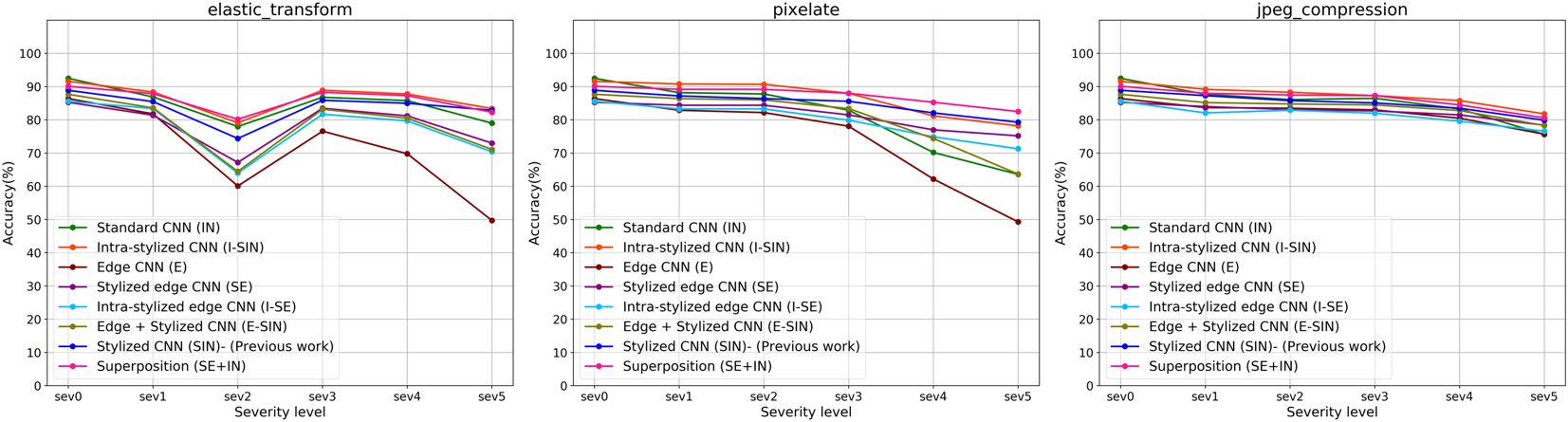}
\end{subfigure}
\caption{Performance of different networks on ImageNet-C corruptions at different severity levels}
\label{fig:appendix_networks_corruptions}
\end{figure}

We show in Section \ref{sec:robustness_corruptions} that networks E, SE, E-SIN performs poorly on corruptions despite having stronger shape bias than SIN. We also show that superposition of SE with natural images IN (SE+IN) slightly outperforms SIN even having lower shape bias respectively. In Figure \ref{fig:appendix_networks_corruptions}, we show that these results are consistent across all 15 ImageNet-C distortions at different severity levels. 

\subsection{Finetuning affine parameters on different corruptions}
\label{sec:appendix_finetune_affine}
\begin{figure}
\centering
\begin{subfigure}[h]{1.0\linewidth}
\includegraphics[width=\linewidth, height=0.19\textheight]{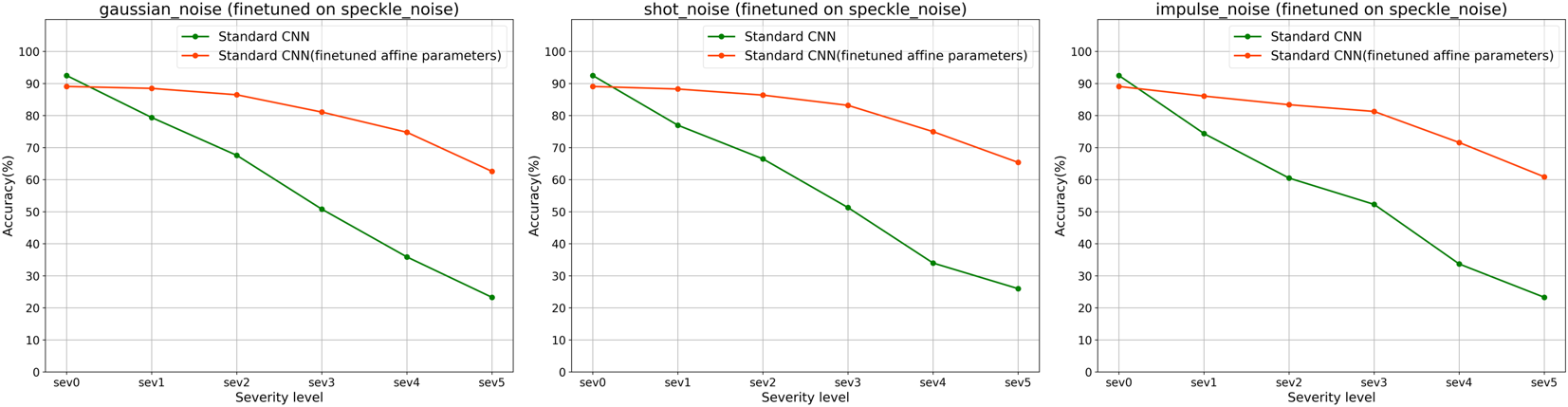}
\end{subfigure}

\begin{subfigure}[h]{1.0\linewidth}
\includegraphics[width=\linewidth, height=0.19\textheight]{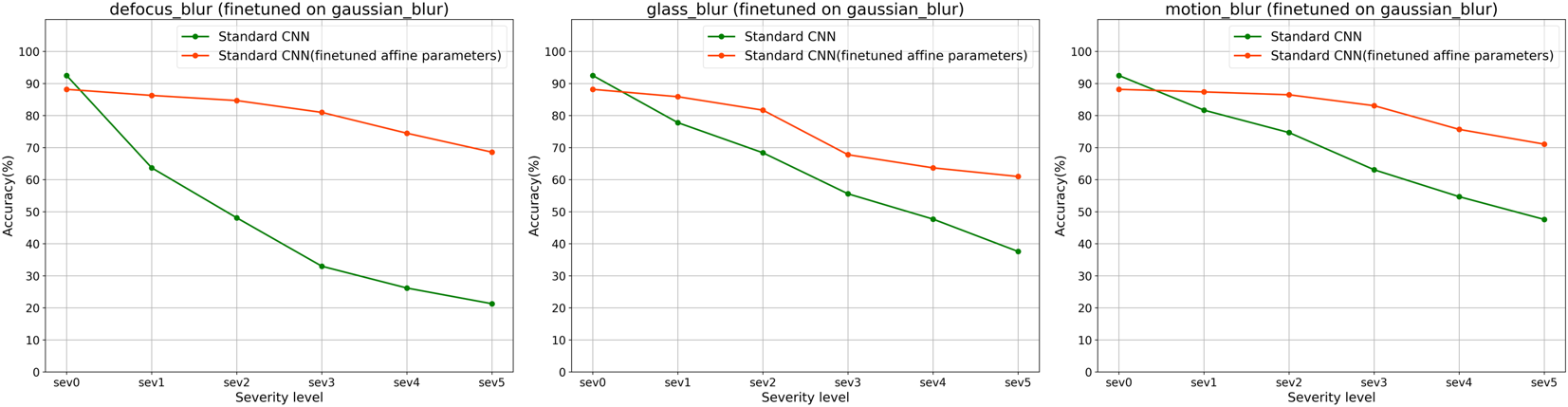}
\end{subfigure}

\begin{subfigure}[h]{0.325\linewidth}
\includegraphics[width=\linewidth, height=0.19\textheight]{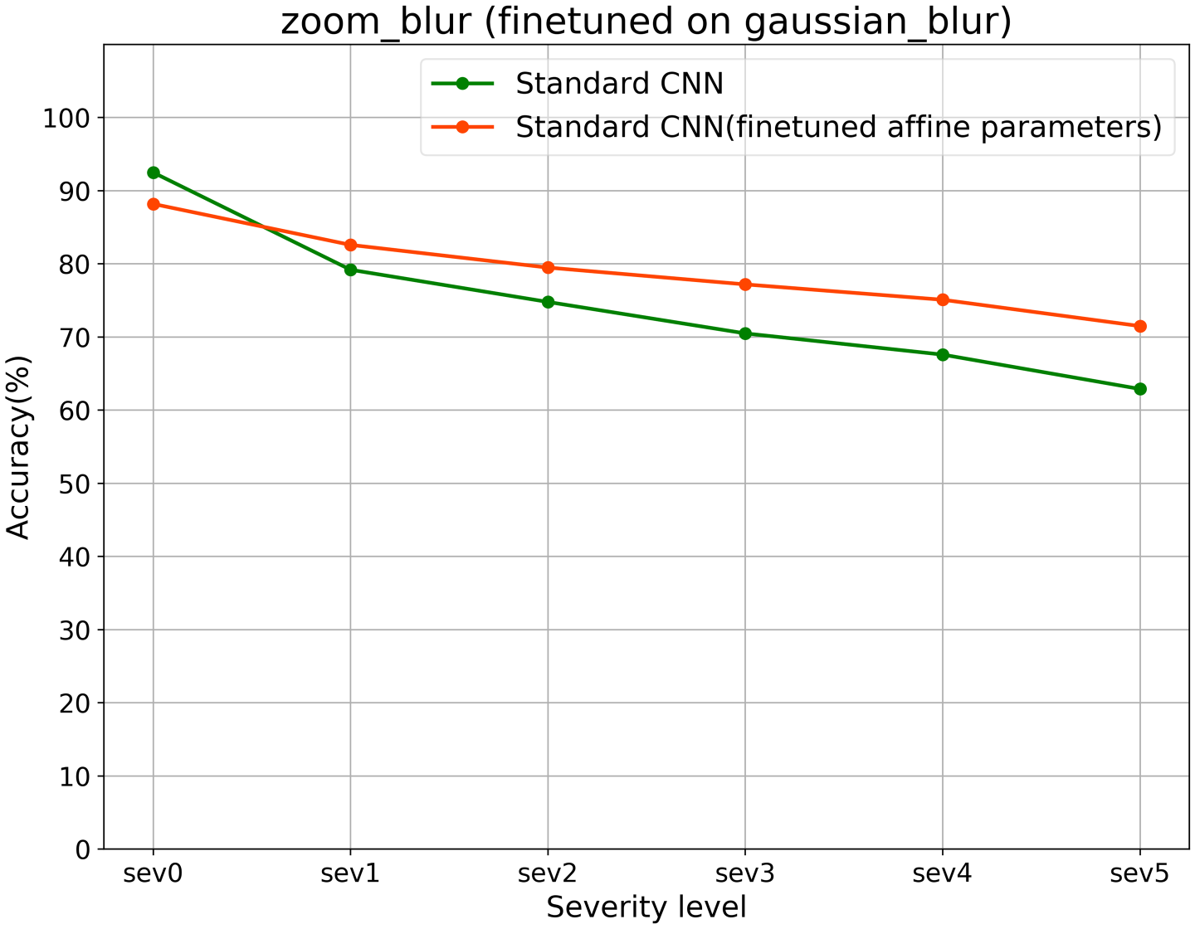}
\end{subfigure}
\begin{subfigure}[h]{0.325\linewidth}
\includegraphics[width=\linewidth, height=0.19\textheight]{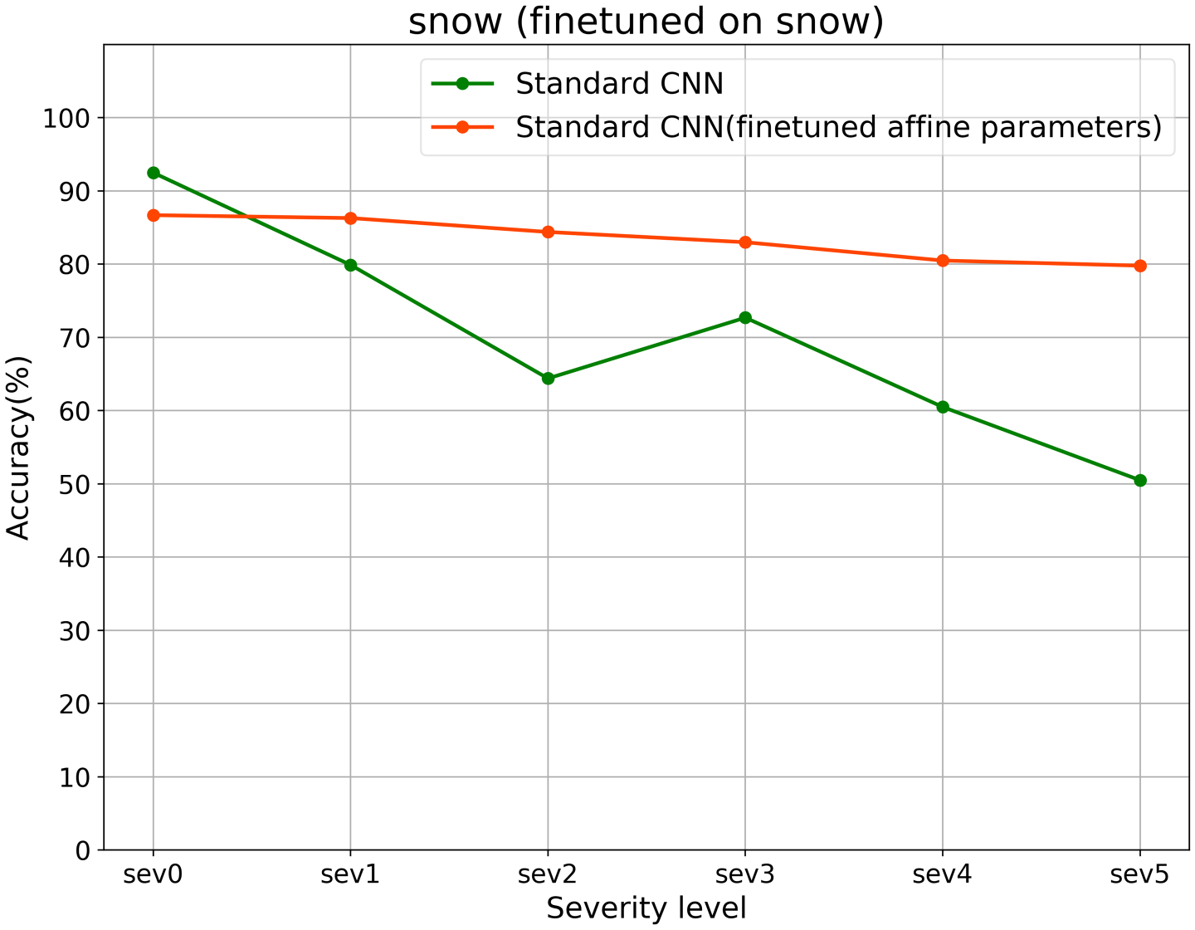}
\end{subfigure}
\begin{subfigure}[h]{0.325\linewidth}
\includegraphics[width=\linewidth, height=0.19\textheight]{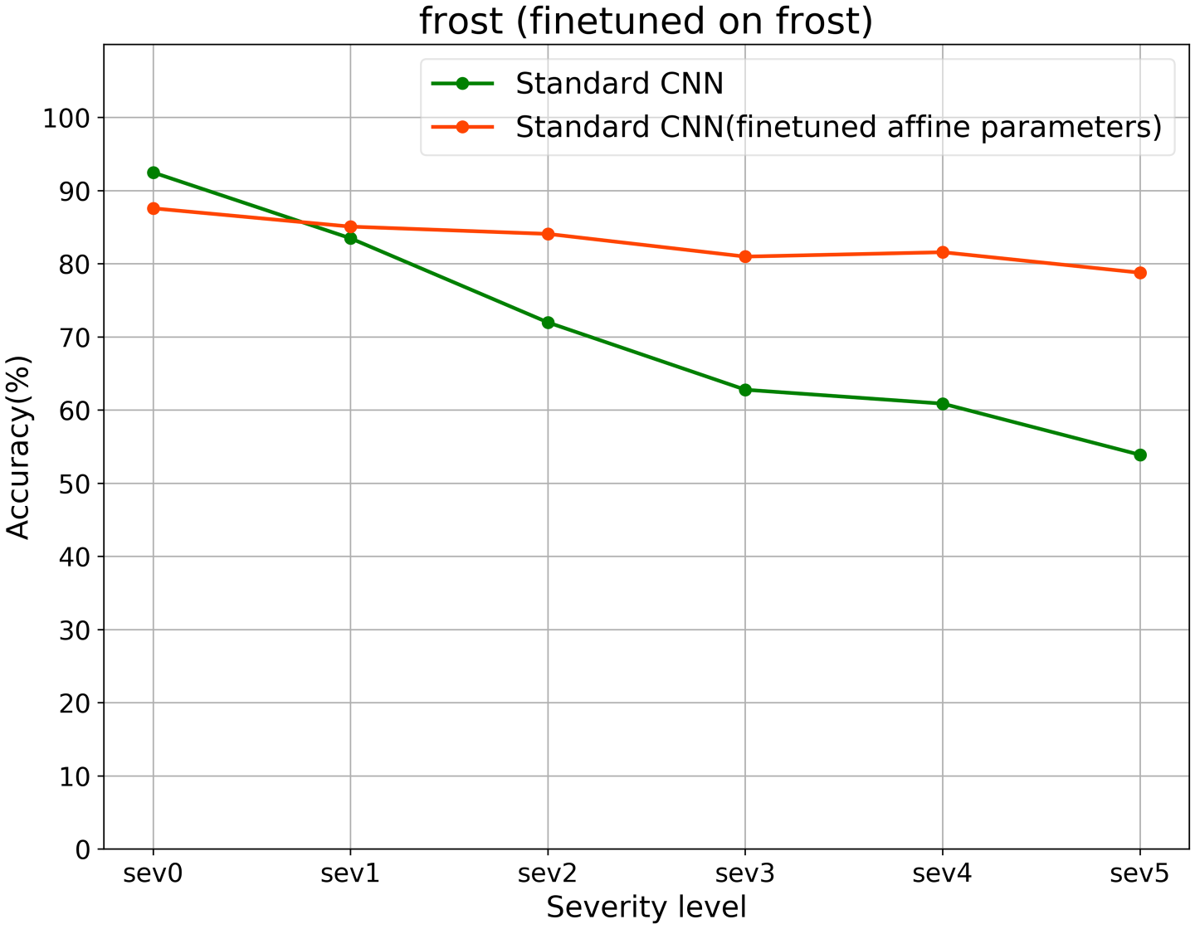}
\end{subfigure}

\begin{subfigure}[h]{0.325\linewidth}
\includegraphics[width=\linewidth, height=0.19\textheight]{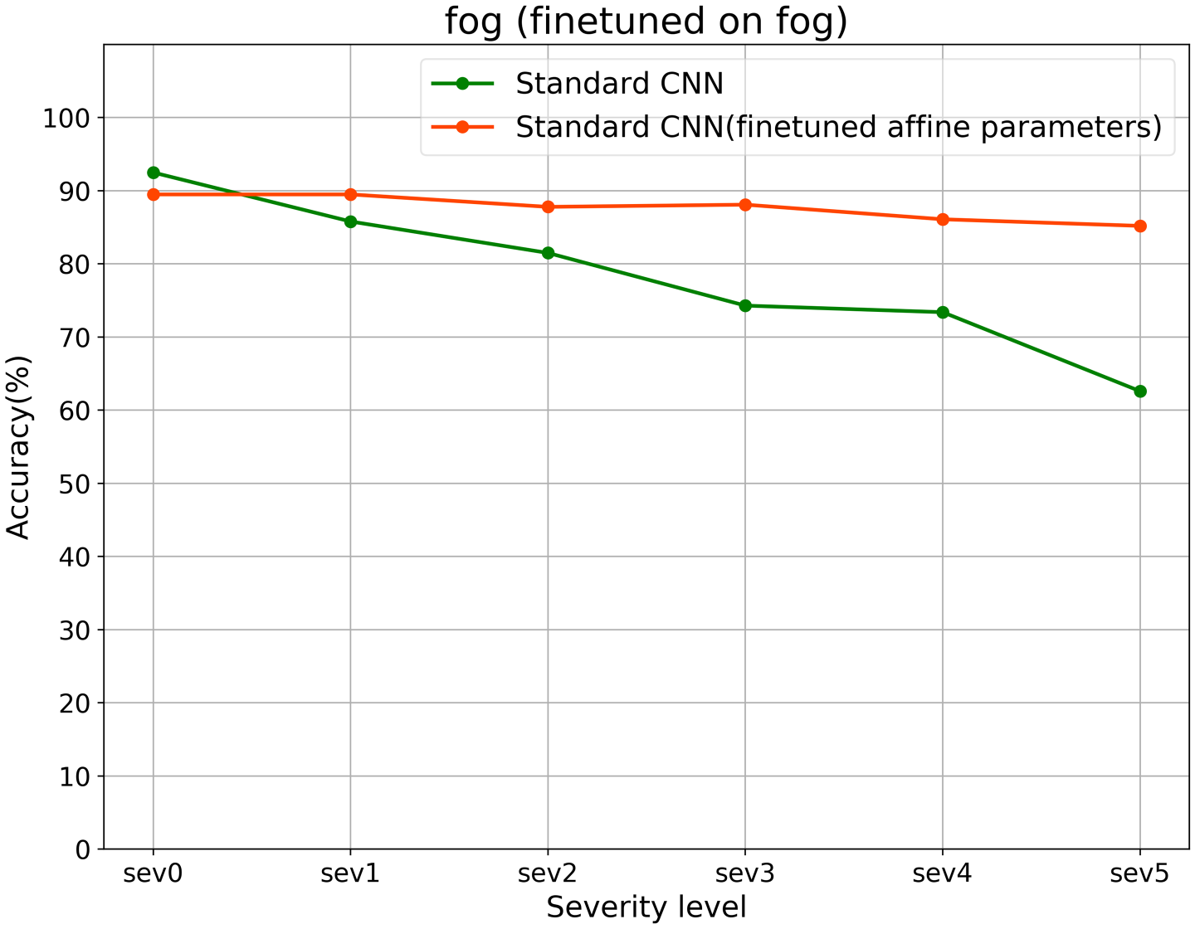}
\end{subfigure}
\begin{subfigure}[h]{0.325\linewidth}
\includegraphics[width=\linewidth, height=0.19\textheight]{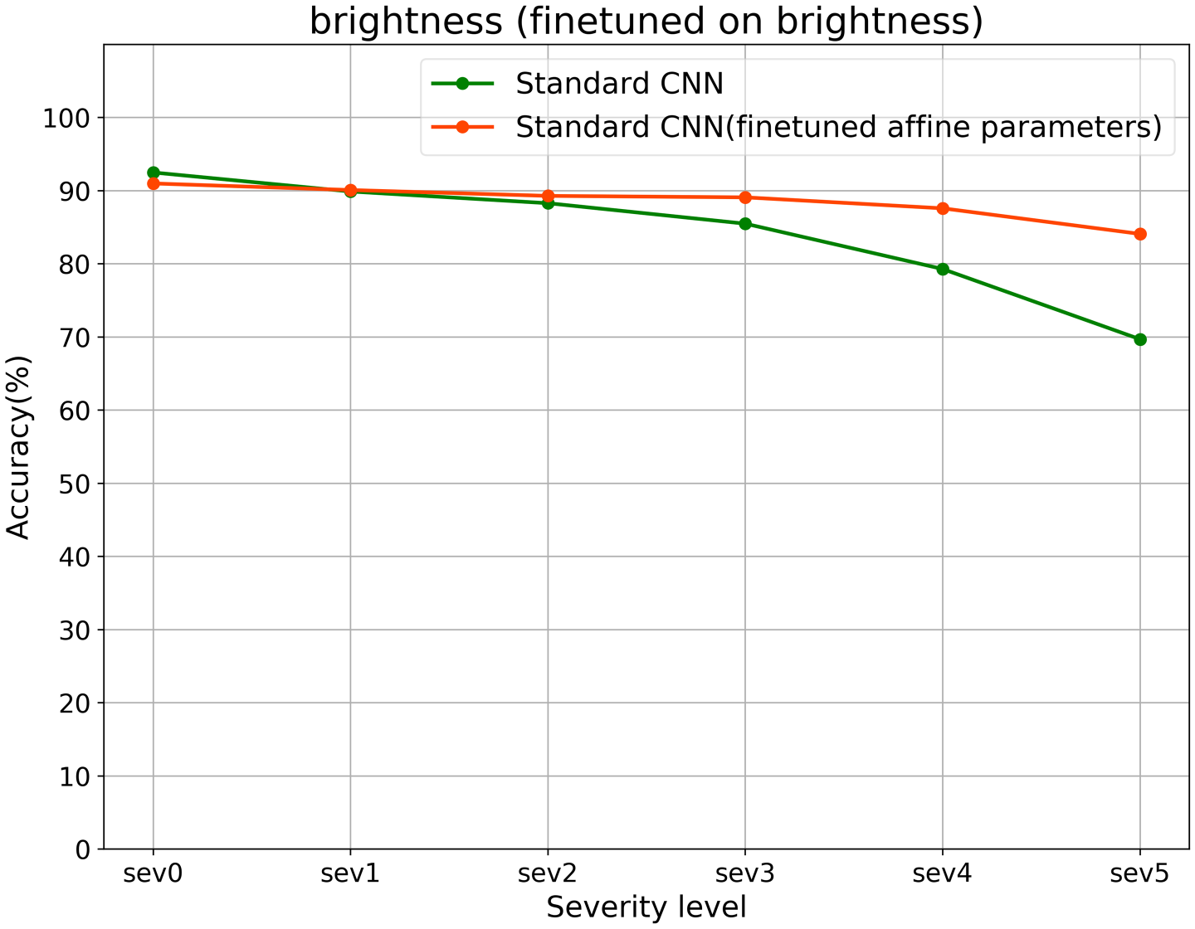}
\end{subfigure}
\begin{subfigure}[h]{0.325\linewidth}
\includegraphics[width=\linewidth, height=0.19\textheight]{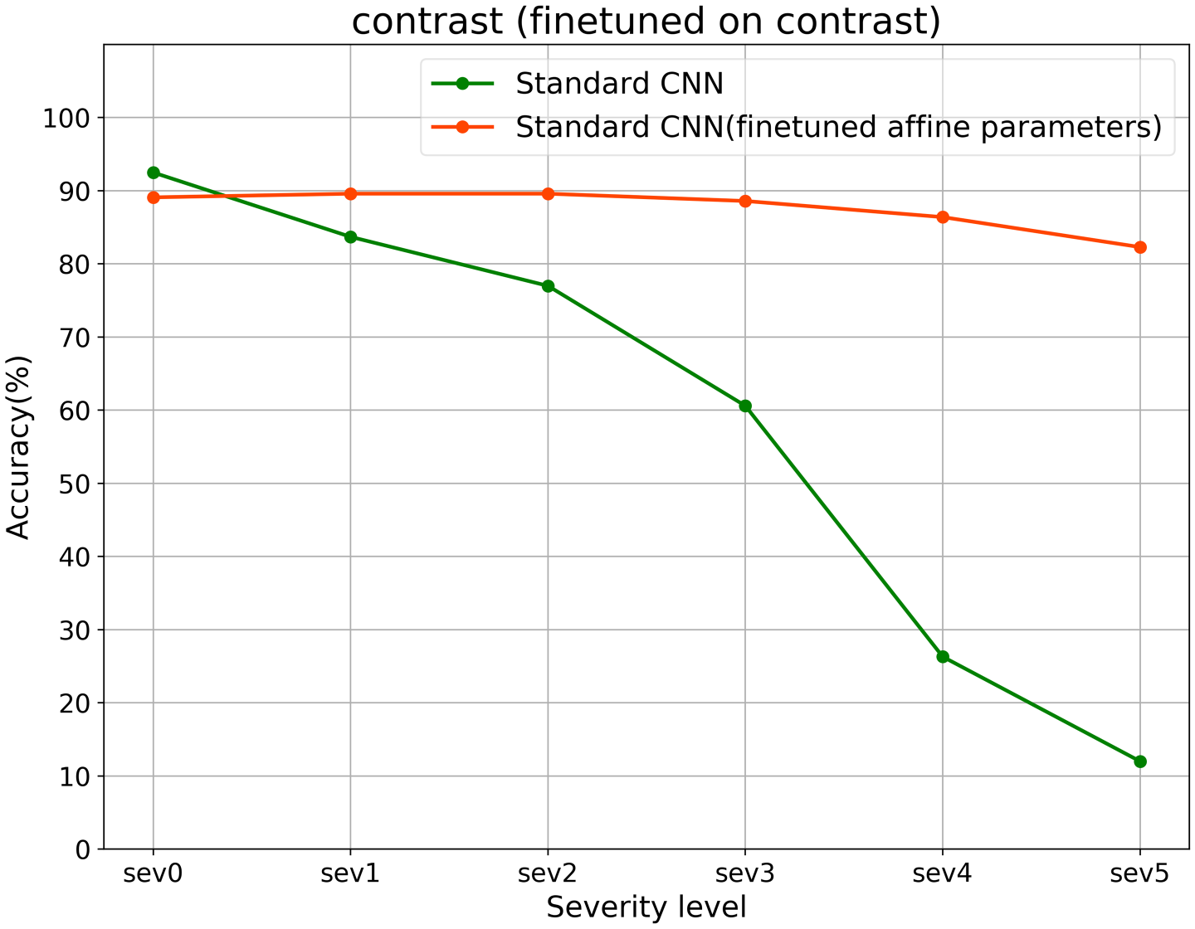}
\end{subfigure}

\begin{subfigure}[h]{0.325\linewidth}
\includegraphics[width=\linewidth, height=0.19\textheight]{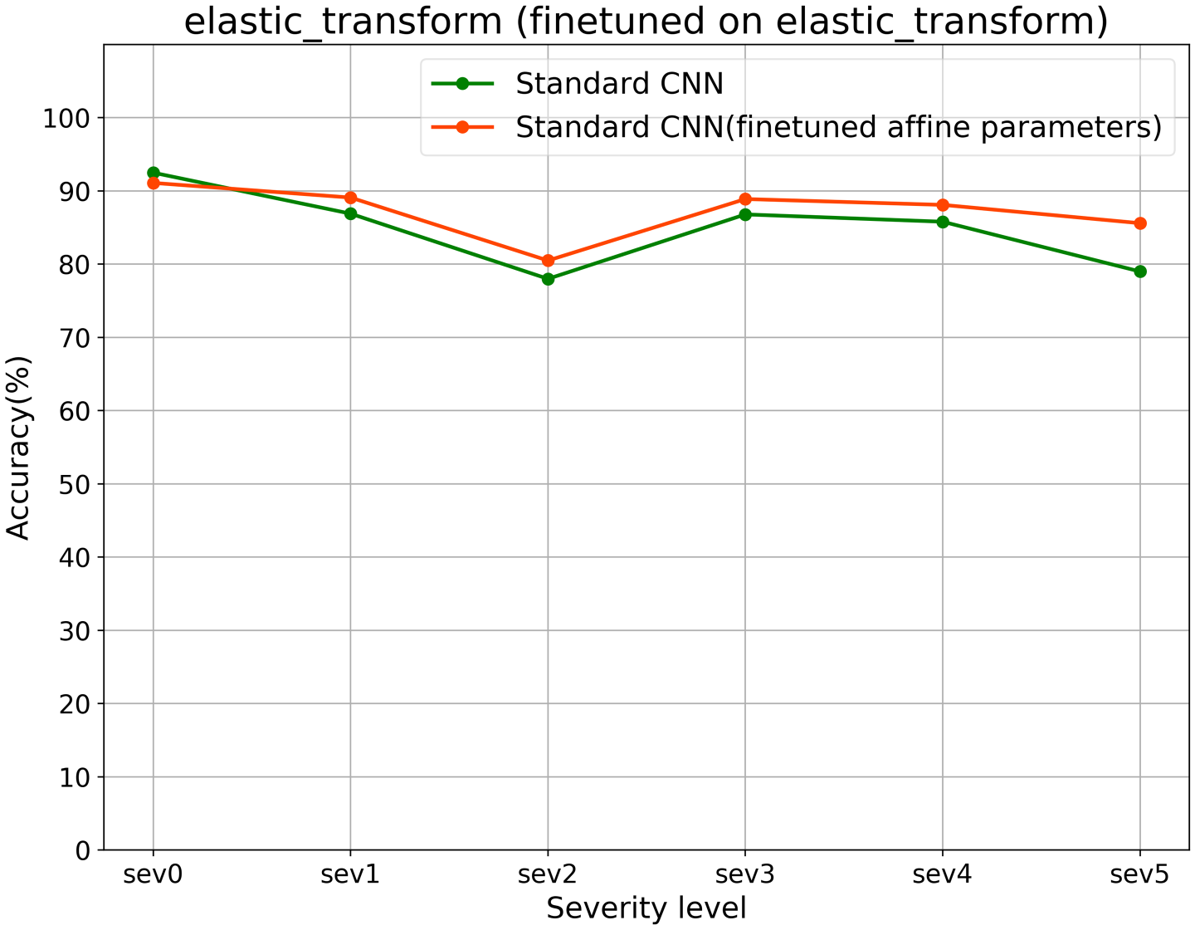}
\end{subfigure}
\begin{subfigure}[h]{0.325\linewidth}
\includegraphics[width=\linewidth, height=0.19\textheight]{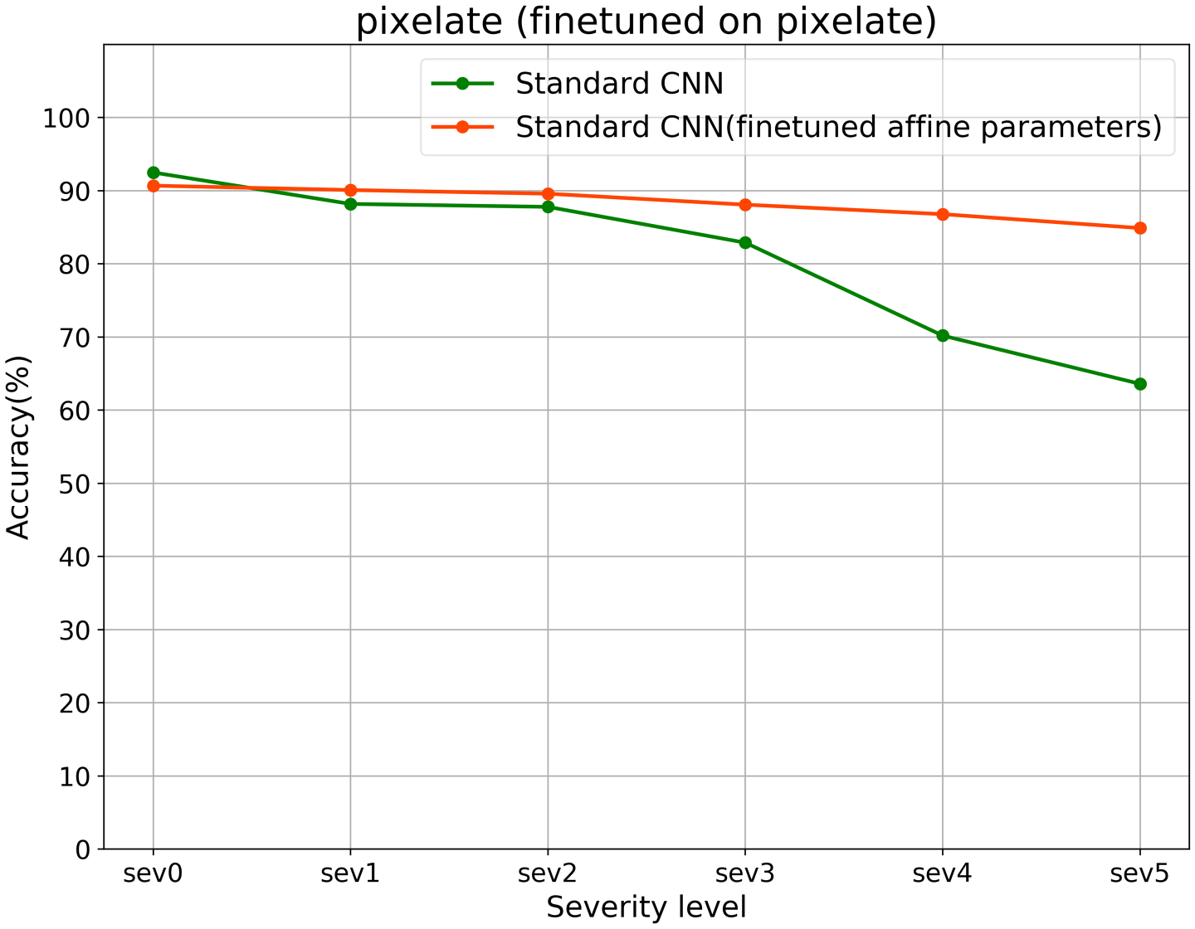}
\end{subfigure}
\begin{subfigure}[h]{0.325\linewidth}
\includegraphics[width=\linewidth, height=0.19\textheight]{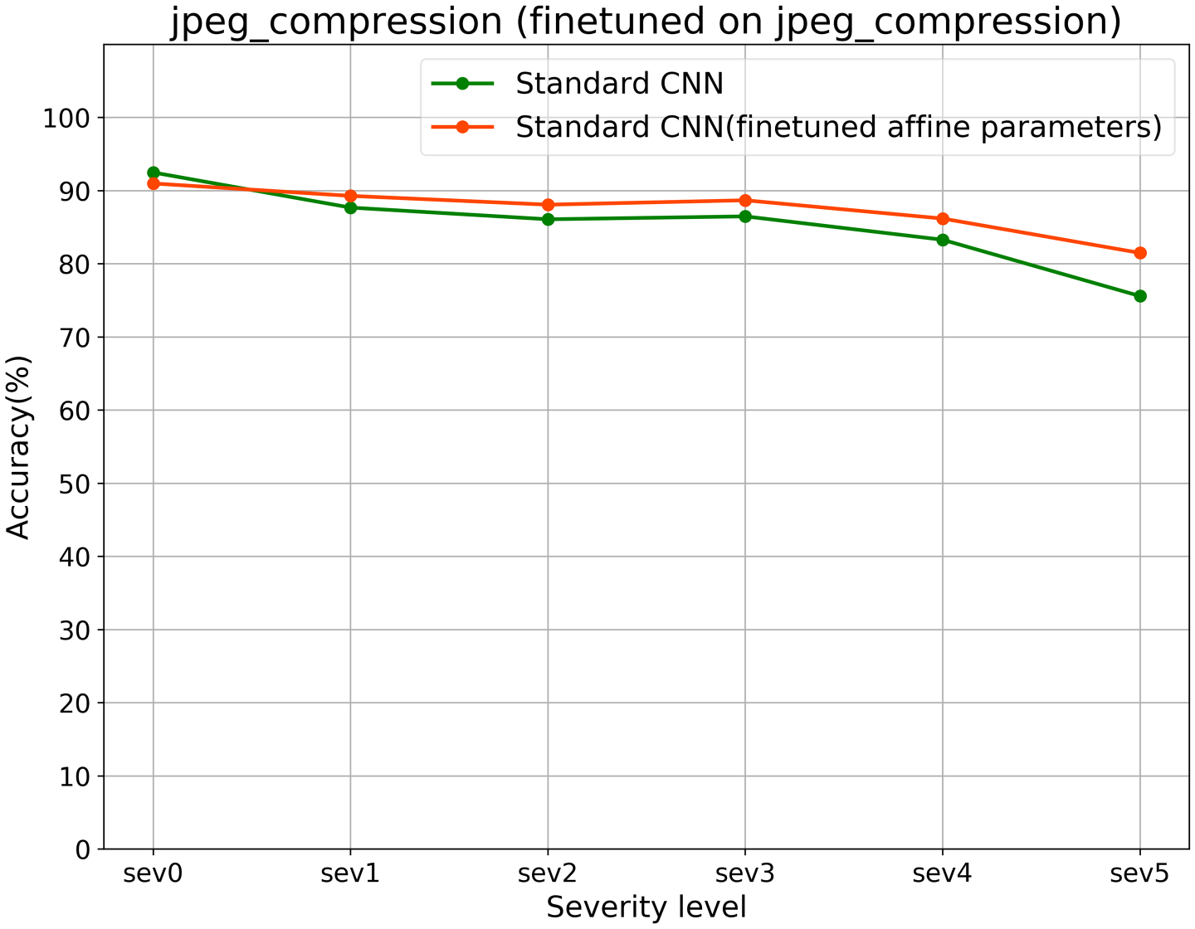}
\end{subfigure}

\caption{Performance of standard network IN on ImageNet-C corruptions when finetuned the affine parameters on the same corruptions}
\label{fig:appendix_corruptions_finetuned}
\end{figure}

As mentioned in Section \ref{sec:finetune_affine}, affine parameters of normalization layers in pre-trained IN are fine-tuned on corruptions from ImageNet-C separately. Here, fine-tuning only the affine parameters of IN on a respective corruption greatly improves the mean corruption accuracy on the same corruption across different severity levels. The affine parameters are fine-tuned on \texttt{speckle noise}, \texttt{gaussian blur}, \texttt{snow}, \texttt{frost}, \texttt{fog}, \texttt{brightness}, \texttt{contrast}, \texttt{elastic transform}, \texttt{pixelate}, \texttt{jpeg compression} separately and resulting 10 different IN networks. Each of these fine-tuned networks are evaluated on the same corruption or similar category of corruptions. For e.g, a network fine-tuned on \texttt{frost} is evaluated only on \texttt{frost} and network fine-tuned on \texttt{speckle noise} is also evaluated on the other set of noises like \texttt{gaussian noise}, \texttt{shot noise}, \texttt{impulse noise}. Note that training data of ImageNet20 is augmented with respective corruptions to fine-tune the affine parameters. Each training sample in a mini-batch is augmented with the corresponding corruption at a randomly chosen severity level. The severity parameters that are already pre-defined for every severity level in ImageNet-C are used. A total of 50 epochs are used for fine-tuning the affine parameters on a corruption, starting with learning rate 0.01 and reduce it to 0.001 after 45 epochs. Performance of the networks that are fine-tuned and evaluated on the same or similar category of corruptions at different severity levels are presented in Figure \ref{fig:appendix_corruptions_finetuned}. As shown in the figure, adapting just the distribution on the learned representations from the standard ImageNet20 is sufficient to achieve high performance on respective distribution of corruptions.

\end{document}